\documentclass[11pt]{article}

\usepackage{acl}

\usepackage{times}
\usepackage{latexsym}
\usepackage{float}
\usepackage{caption}
\usepackage{subcaption}
\usepackage{enumitem}

\usepackage{tikz}
\usetikzlibrary{arrows.meta, positioning}

\newsavebox{\upperhalftribox}
\savebox{\upperhalftribox}{\tikz[baseline=0.0ex]{\draw[black, line width=0.5pt] (0,0.28) -- (0.28,0.28) -- (0,0) -- cycle;}}
\newcommand{\upperhalftri}{\usebox{\upperhalftribox}}

\newsavebox{\lowerhalftribox}
\savebox{\lowerhalftribox}{\tikz[baseline=0.0ex]{\draw[black, line width=0.5pt] (0.28,0) -- (0.28,0.28) -- (0,0) -- cycle;}}
\newcommand{\lowerhalftri}{\usebox{\lowerhalftribox}}

\usepackage{capt-of}

\usepackage{amsmath}
\usepackage{microtype}
\usepackage{wrapfig}
\usepackage{inconsolata}
\usepackage{graphicx}
\usepackage{comment}

\usepackage{tabularx}
\usepackage{booktabs}

\usepackage[space]{xeCJK}
\usepackage{polyglossia}
\setmainlanguage{english}

\usepackage{longtable}
\usepackage{array}
\usepackage{cleveref}

\setotherlanguage{hindi}
\newfontfamily\devanagarifont[
    Script=Devanagari
]{FreeSerif.otf}

\graphicspath{{img/}}

\renewcommand{\arraystretch}{1.15}

\usepackage{stfloats}

\title{Beyond Factual QA: Mentorship-Oriented Question Answering over\\ Long-Form Multilingual Content}

\author{
Parth Bhalerao \quad
Ruiwen Guan \quad
Diola Dsouza \quad
Oana Ignat \\
Department of Computer Science and Engineering, Santa Clara University \\
Santa Clara, CA, USA \\
\texttt{\{pbhalerao,oignat\}@scu.edu}
}

\begin{document}
\maketitle

\begin{abstract}
Question answering systems are usually evaluated for factual correctness, but many real world settings such as education and career guidance require mentorship responses that offer reflection, context, and guidance. We introduce \textsc{MentorQA}, the first multilingual dataset and evaluation framework for mentorship focused question answering from long form videos, with nearly 9,000 QA pairs from 180 hours of content across four languages. We define evaluation dimensions that measure fluency, clarity, alignment, and mentorship value, and use them to compare Single Agent, Dual Agent, RAG, and Multi Agent QA pipelines under controlled conditions. Multi Agent pipelines produce the strongest mentorship responses, with larger gains for complex topics and lower resource languages. We also study LLM based evaluation and find substantial variation in alignment with human judgments. Overall, \textsc{MentorQA} establishes mentorship focused QA as a distinct research problem and provides a benchmark for studying agentic architectures and evaluation design in educational AI. The dataset and evaluation framework are released at \url{https://aim-scu.github.io/MentorQA}.
\end{abstract}

\begin{table}[!t]
\centering
\footnotesize
\setlength{\tabcolsep}{3pt}
\renewcommand{\arraystretch}{1.02}
\begin{tabular}{p{0.96\columnwidth}}
\toprule

\textbf{English \;|\; Career/Education \;|\; Multi-Agent} \\
\textbf{Q:} What should a student consider when choosing a research topic that aligns with their interests? \\
\textbf{A:} When selecting a research topic that aligns with their interests, a student should consider several key factors: 1. **Interest and Passion**: The student must genuinely... \\
\textbf{Scores:} QFlu~5, AFlu~5, QClar~4, AClar~5, QA-Algn~5, QMen~5, AMen~5 \\
\midrule

\textbf{Chinese \;|\; Education \;|\; Single-Agent} \\
\textbf{Q:} 千禧世代的出生年份是什麼？ 
\textit{(What are the birth years of the Millennial generation?)} \\
\textbf{A:} 千禧世代的出生年份是1980年至1994年或1995年。 \\
\textit{(The birth years of the Millennial generation are 1980 to 1994 or 1995.)} \\
\textbf{Scores:} QFlu~3, AFlu~4, QClar~5, AClar~2, QA-Algn~4, QMen~1, AMen~1 \\
\midrule

\textbf{Hindi \;|\; Business \;|\; Dual-Agent} \\
\textbf{Q:}
\begin{hindi}
भारत की ऐतिहासिक रूप से धनवान था तो उसके व्यापार का क्या रूप था?
\end{hindi} 
\textit{(If India was historically wealthy... how was the nature of trade?)} \\
\textbf{A:}
\begin{hindi}
भारत ने चानक्य के समय से बीटा को बोर्ड ट्रेड के एक अच्छे देश के रूप में जाना था।..... डाट बिक था जो तब विश्व की दुनिया का नक्षा था।
\end{hindi} \\
\textit{(From Chanakya's time, India was known as.. and carried out trade with..)} \\
\textbf{Scores:} QFlu~3, AFlu~3, QClar~2, AClar~3, QA-Algn~5, QMen~1, AMen~3 \\
\midrule

\textbf{Romanian \;|\; Mental Health \;|\; RAG} \\
\textbf{Q:} Ce implică muncă prea puțin stimulantă și ce efecte are asupra sistemului nervos?
\textit{(What does work that is too little stimulating imply..... and affects nervous system?)} \\
\textbf{A:} Muncă prea puțin stimulantă implică simțirea de plictisire și blazare, ceea ce poate duce la o solicitare excesivă... \\
\textit{(Work that is not stimulating enough involves feelings of boredom.... lead to excessive strain on the nervous system, causing......)}\\
\textbf{Scores:} QFlu~2, AFlu~4, QClar~2, AClar~4, QA-Algn~1, QMen~4, AMen~4 \\
\bottomrule
\end{tabular}
\caption{\textsc{MentorQA} multilingual examples across languages, topics, and generation models.
Scores denote Fluency (Q/AFlu), Clarity (Q/AClar), QA Alignment (QA-Algn), and Mentorship (Q/AMen) on a 1 to 5 Likert scale.}
\vspace{-2em}
\label{tab:dataset}
\end{table}

\section{Introduction}
\label{sec:introduction}

Question answering (QA) systems are typically evaluated on factual correctness~\cite{mulla2023automatic, fu-etal-2024-qgeval}. However, many real world settings such as education, career guidance, and personal development require \emph{mentorship}: answers that provide context, reflection, and guidance rather than isolated facts~\cite{ma2019automatic, kovacs2016effects}. Existing QA benchmarks rarely capture this need, especially for multilingual and long form content where useful guidance is distributed across extended narratives.

This gap limits the study of mentorship focused QA in two ways. First, prior QA research has largely emphasized short form inputs, factual accuracy, and monolingual evaluation~\cite{alfaraby2023review}. Multilingual QA benchmarks assess fluency and correctness, but seldom measure whether answers provide learning value or actionable guidance. Second, while recent language models have advanced open domain QA~\cite{guo2024survey}, it remains unclear how architectural choices such as retrieval augmentation and agentic coordination affect mentorship quality across languages with different discourse norms and resource availability.

\paragraph{Contributions.} \textbf{First, we introduce \textsc{MentorQA}, the first multilingual dataset and evaluation framework for mentorship focused question answering from long form videos.} The dataset contains nearly 9,000 QA pairs from 180 hours of mentorship content across four languages. 
\textbf{Second, we define evaluation dimensions that capture clarity, alignment, and mentorship value, and compare Single Agent, Dual Agent, RAG, and Multi Agent QA architectures under controlled settings.} Our results show that Multi Agent pipelines consistently achieve higher mentorship quality, with particularly strong gains for complex topics and lower resource languages.
\textbf{Finally, we examine the reliability of automated LLM based evaluation across languages and metrics}, finding substantial variation in alignment with human judgments. 
Throughout this work, mentorship focused QA refers to the task objective, while mentorship value refers to the learning value and actionable guidance captured by our Question Mentorship and Answer Mentorship metrics.
\textbf{Together, our findings establish mentorship focused QA as a distinct research problem and provide a multilingual benchmark for studying agentic architectures and evaluation methods in educational AI.}

\section{Related Work}

\paragraph{Educational and Mentorship QA from Long Transcripts.}
Recent work extracts questions and answers from long educational recordings through lecture summaries~\cite{chen2024e-qgen}, video based QA context selection~\cite{yu2025context}, and segment level retrieval for podcasts and meeting transcripts~\cite{elaryan2022, prasad2023meetingqa}. Other work studies personalized QA for language learning assistants~\cite{sammoudi2025qa} and multi agent debate for online education~\cite{du2025enhancing}. These systems support educational content, but they mainly optimize factual correctness, content coverage, or single modeling settings. In contrast, mentorship focused QA emphasizes guidance, reflection, and learning value in long form multilingual content, which remains underexplored.

\paragraph{Agentic and RAG Based QA Pipelines.}
Agentic models support complex workflows for long document understanding and QA by coordinating specialized agents for reading, reasoning, and generation~\cite{wang2025multiagent, saadaoui2025coordinated}, with recent surveys summarizing common collaboration mechanisms~\cite{tran2025multiagentcollab}. Retrieval augmented generation has also been used for domain specific QA, including health risk assessment~\cite{meng2025using} and podcast or meeting analysis~\cite{aquilina2023end, zhu2025-m3finmeeting}. However, these approaches are rarely compared under controlled conditions or evaluated for mentorship quality. We compare Single Agent, Dual Agent, RAG, and Multi Agent pipelines with a fixed base model to isolate the effect of architecture on mentorship focused QA.

\paragraph{Multilingual QA and Metric Based Evaluation.}
Prior work has studied question generation evaluation and multilingual QA through automatic QG metrics~\cite{khapra-2018-towards}, reference free and answer based metrics such as RQUGE~\cite{mohammadshahi-etal-2023-rquge} and QG Eval~\cite{fu-etal-2024-qgeval}, LLM as judge evaluation~\cite{wolfe2024llm-as-a-judge}, and multilingual QA benchmarks~\cite{ushio-etal-2023-practical, moreno-cediel2024evaluating, ruder-sil-2021-multi, asai2021one}. Yet automatic metrics often correlate weakly with human judgments for task oriented dimensions such as usefulness or guidance~\cite{fu-etal-2024-qgeval}. We therefore introduce mentorship focused metrics that capture learning value and support comparison between human and automatic evaluation.

\section{The \textsc{MentorQA} Dataset}

We introduce \textsc{MentorQA}, to our knowledge the first QA dataset focused on mentorship. Unlike factual QA benchmarks, \textsc{MentorQA} centers on guidance, reflection, and practical insight. We define mentorship value as information that helps learners grow through advice, perspective, or actionable guidance. The dataset contains 8,990 QA pairs from 120 long form mentorship videos across four languages and six topics.

\noindent\textbf{Video Collection and Pre Processing.}
We collect podcast style conversations and panel discussions from YouTube that provide reflective, experience driven guidance. Each video averages over 1.5 hours, resulting in 180 hours of multilingual mentorship content. We process only the audio track and transcribe it with \href{https://github.com/openai/whisper}{Whisper}, which detects the source language and produces transcripts for QA generation.

\noindent\textbf{Languages.}
The dataset covers \textit{English}, \textit{Chinese}, \textit{Hindi}, and \textit{Romanian}, selected for typological and cultural diversity while supporting reliable annotation. Videos and QA pairs are balanced across languages and evaluated by native or expert speakers.

\noindent\textbf{Topic Selection across Languages.}
Videos cover six mentorship topics\footnote{All the mentorship videos are in \Cref{sec:videos}.}: \textit{Entrepreneurship, Education, Finance, Mental Health, Personal Growth}, and \textit{Physical Health}. Topic labels are assigned using \textit{Qwen2.5 7B}~\cite{yang2025qwen2.5-1m} and verified by expert annotators. English videos focus on academic and professional mentorship, including research directions, publishing, and PhD pathways through \href{https://www.youtube.com/@aclmentorship}{ACL Mentorship}. Romanian videos cover education, personal development, and mental health through conversations with authors, coaches, and athletes from \href{https://www.youtube.com/@MindArchitect}{Mind Architect}. Chinese videos focus on experience driven mentorship in mental health, career growth, finance, and physical well being from \href{https://www.youtube.com/@DrinkingLibrary}{Drinking Library}. Hindi videos focus on career advancement, business, and mental or physical health from \href{https://www.youtube.com/@ranveerallahbadia}{The Ranveer Show}.

\noindent\textbf{QA Generation.}
We generate QA pairs from each transcript using \textit{Single Agent}, \textit{Dual Agent}, \textit{RAG}, and \textit{Multi Agent}. Each model produces about 20 mentorship focused QA pairs per video, giving about 80 QA pairs per video and 8,990 pairs overall (\autoref{tab:stats}). Multilingual examples are shown in \Cref{tab:dataset}, and model architectures are described in \Cref{sec:method}.

\begin{table}[h]
\centering
\footnotesize
\setlength{\tabcolsep}{6pt}
\renewcommand{\arraystretch}{1.1}
\begin{tabular}{l r}
\toprule
\multicolumn{2}{l}{\textbf{Dataset Scale and Coverage}} \\
\midrule
Videos (long form) & 120 \\
Total duration (hours) & 180 \\
Languages & 4 \\
Mentorship topics & 6 \\
QA generation models & 4 \\
Total QA pairs & 8,990 \\
\midrule
\multicolumn{2}{l}{\textbf{Evaluation and Quality Control}} \\
\midrule
Evaluation dimensions & 4 \\
LLM judges & 9 \\
Human annotators & 12 \\
\midrule
\multicolumn{2}{l}{\textbf{Mentorship Topics (\# QA pairs)}} \\
\midrule
Personal Growth & 4,450 \\
Career \& Education & 2,689 \\
Mental Health & 741 \\
Business \& Entrepreneurship & 392 \\
Physical Health & 431 \\
Finance & 287 \\
\bottomrule
\end{tabular}

\caption{\textsc{MentorQA} statistics.}
\label{tab:stats}
\vspace{-1.5em}
\end{table}

\noindent\textbf{Quality Assurance.}
Each QA pair is evaluated along four dimensions: \textit{Fluency}, \textit{Clarity}, \textit{QA Alignment}, and \textit{Mentorship Value}. Human evaluation uses three annotators per language, complemented by nine LLM based judges for scalable analysis and comparison with human judgments. Evaluation details are in \Cref{sec:evaluation}.

\noindent\textbf{PII Anonymization.}
Although the source data are public, we reduce downstream propagation of personally identifiable information. We combine named entity recognition based on \href{https://huggingface.co/Davlan/xlm-roberta-base-ner-hrl}{XLM RoBERTa}, structured PII detection and anonymization via \href{https://github.com/microsoft/presidio}{Presidio}, LLM prompts for implicit identifiers, and human review for lower resource languages and culturally specific entities.

\noindent\textbf{Scalability.}
\textsc{MentorQA} is released as an extensible open source resource. Although the current version covers four languages, selected channels, and one video platform, the codebase and documentation support expansion to additional languages and content sources.

\section{Mentorship focused QA Generation}\label{sec:method}

Mentorship focused QA from long form conversational videos requires more than long context reasoning and topic tracking. Mentorship value is unevenly distributed across transcripts: brief segments with reflection, advice, or experience driven insight can be more useful than longer descriptive passages. Effective systems must therefore identify valuable guidance and allocate questions by mentorship quality rather than transcript length.

We study four QA generation pipelines of increasing structure: \textit{Single Agent}, \textit{Dual Agent}, \textit{Retrieval Augmented Generation (RAG)}, and \textit{Multi Agent}. These pipelines introduce topic awareness, answer grounding, and explicit control over mentorship quality and question allocation. Across all pipelines, we use the same base model, \textit{Qwen 2.5 7B Instruct 1M}~\cite{yang2025qwen2.5-1m}, in long context mode, ensuring that differences reflect architecture rather than model capacity. Full prompts are provided in \autoref{sec:prompts}.

The four pipelines are designed as distinct architectures to isolate architectural effects. Topic aware chunking, retrieval based grounding, and question scoring with filtering represent separate design dimensions; combining them would obscure which component drives performance differences.

\subsection{Single Agent}\label{sec:single}
The Single Agent baseline processes the full transcript with one long context LLM. The model is prompted as an expert educational analyst to generate 20 mentorship focused QA pairs that cover the conversation, avoid trivial or overly local questions, and provide educational value. This baseline relies only on the long context capability of the LLM, without segmentation, retrieval, or coordination.

\subsection{Dual Agent}
The Dual Agent pipeline separates topic segmentation from QA generation to reduce over concentration in dense or early transcript regions. A \textit{Chunking Agent} partitions the transcript into coherent topic based segments while preserving the original content and adapting segment length to topic duration. A \textit{QA Agent} then generates a fixed number of mentorship focused QA pairs per segment using the same prompt as the Single Agent baseline. This improves coverage across the conversation and better aligns questions with mentorship themes.

\subsection{Retrieval Augmented Generation (RAG)}
RAG separates question generation from answer grounding. The model first generates a fixed set of diverse mentorship focused questions from the full transcript. Each question is then answered using relevant transcript segments retrieved through similarity search. We build a multilingual retrieval index with BGE M3~\cite{chen-etal-2024-m3}, which chunks long transcripts and supports all four dataset languages. Retrieved chunks are provided to a second LLM query to generate the final answer, enabling global question coverage while grounding answers in relevant transcript evidence.

\subsection{Multi Agent}
\begin{figure}[h]
\centering
\includegraphics[width=1.0\columnwidth]{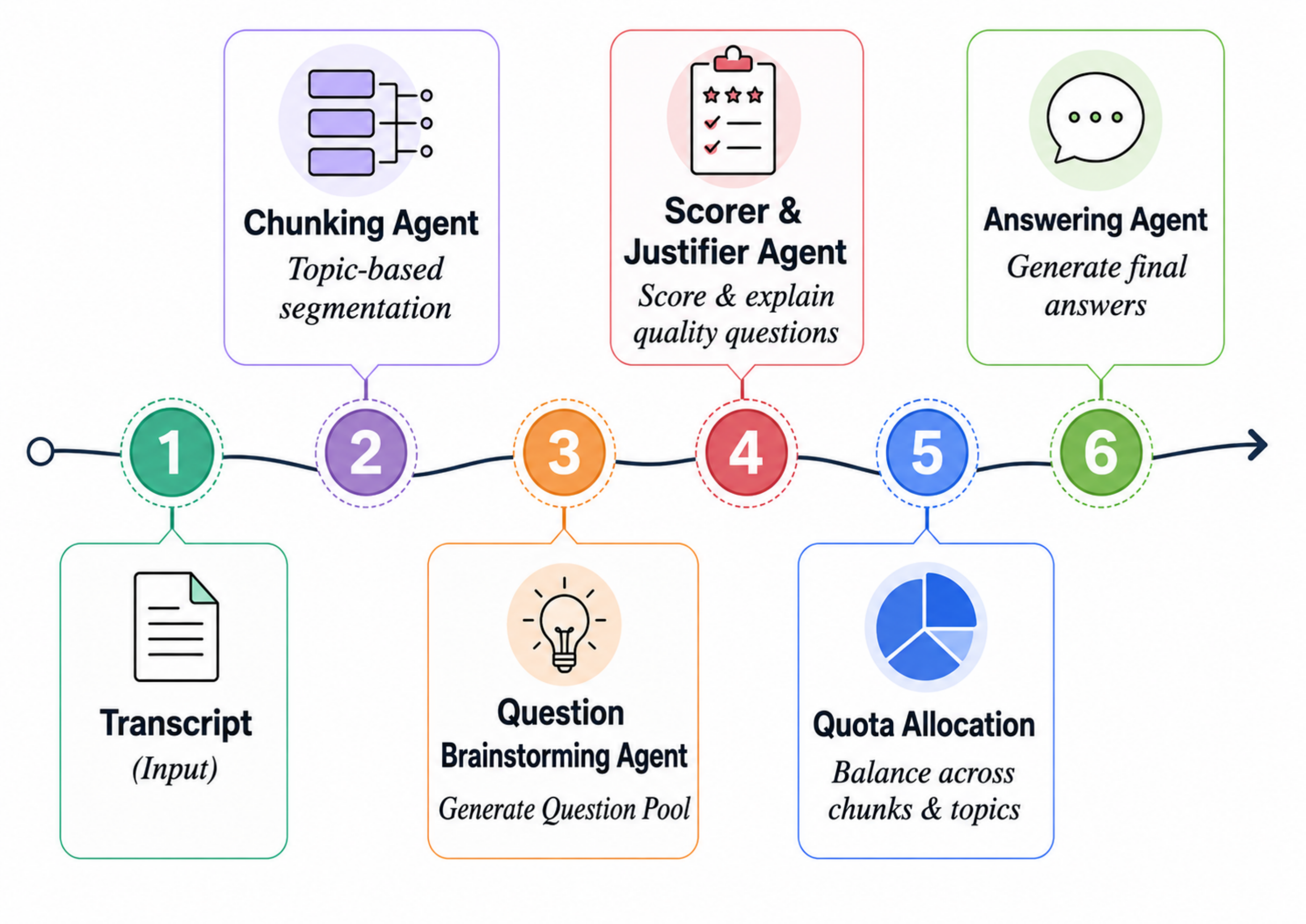}
\caption{Multi Agent QA generation pipeline with topic segmentation, question exploration, quality scoring, quota based allocation, and grounded answer generation.}
\label{fig:multiagent}
\vspace{-1.5em}
\end{figure}

Dual Agent improves coverage through topic based segmentation, and RAG grounds answers through retrieval, but both use single pass question generation.
\textit{In practice, mentorship quality varies substantially across topics: uniform question allocation can over represent weak segments and under represent parts of the conversation rich in guidance.}
We therefore introduce a Multi Agent framework that scores mentorship quality and allocates questions across topics based on estimated value.
As shown in \autoref{fig:multiagent}, the pipeline uses specialized agents for topic discovery, question exploration, quality assessment, allocation, and answer generation.

The process starts with the same \textit{Chunking Agent} used in Dual Agent, which performs dynamic topic based segmentation and assigns each segment a concise title.
For each segment, a \textit{Question Brainstorming Agent} over generates diverse mentorship relevant candidate questions.
A \textit{Scorer Agent} assigns each candidate question $q$ an integer mentorship score, $score(q)\in\{1,\ldots,10\}$.
We then compute segment strength as:
\begin{equation}
    S(i) = \mu(q) - \sigma(q),
\end{equation}
where $\mu(q)$ and $\sigma(q)$ denote the mean and standard deviation of question scores within segment $i$.
This favors segments with consistently high mentorship value and penalizes noisy or uneven question quality.
Given a target of $T=20$ questions per video, matching the generation budget used by the other architectures, we distribute this budget across segments proportionally to their strength scores. Specifically, the quota for segment $i$ is:
\begin{equation}
    \text{quota}(i) =
    \text{round}\!\left(
    \frac{S(i)}{\sum_{j \in \mathcal{S}} S(j)} \times T
    \right),
    \label{eq:quota_raw}
\end{equation}
ensuring that stronger segments, those with higher density of mentorship relevant content, contribute more questions to the final set.
Within each segment, only the top ranked questions are retained.
A \textit{Justifier Agent} explains question selection or rejection to improve transparency and reduce redundancy.
Finally, an \textit{Answering Agent} generates grounded final answers from the corresponding segment with mentorship oriented explanations.
Overall, this design enables broader question exploration, explicit mentorship quality control, and adaptive allocation to the most valuable parts of long form conversations.

\subsection{Evaluation Metrics}
We build on QG Eval~\cite{fu-etal-2024-qgeval}, which separates linguistic and task oriented dimensions, and extend it to capture QA alignment and mentorship value. Fluency and Clarity are evaluated separately for questions and answers to support fine grained analysis.

\paragraph{Linguistic Metrics.}
We assess surface level language quality for both questions and answers:

\begin{itemize}[leftmargin=*, itemsep=0pt, topsep=1pt]
\normalfont
    \item \textbf{Fluency}: \textit{The question or answer is grammatically correct and free of language errors.}
    \item \textbf{Clarity}: \textit{The question or answer is easy to understand, specific, and unambiguous.}
\end{itemize}

\paragraph{Task Oriented Metrics.}
We assess whether QA pairs provide alignment and mentorship effectiveness:

\begin{itemize}[leftmargin=*, itemsep=0pt, topsep=1pt]
\normalfont
    \item \textbf{QA Alignment}: \textit{The question and answer pair demonstrates proper alignment, where the answer adequately satisfies what the question asks.}
    \item \textbf{Question Mentorship}: \textit{The question elicits learning, guidance, advice, or insights that would benefit from mentor expertise.}
    \item \textbf{Answer Mentorship}: \textit{The answer provides guidance, wisdom, practical advice, or insights that help the reader learn or grow.}
\end{itemize}

Together, these metrics capture linguistic quality, semantic alignment, and mentorship effectiveness.

\paragraph{Rating Scale.}
All metrics are rated on a 5 point Likert scale from 1, strongly disagree, to 5, strongly agree. Pilot studies showed that the 3 point scale used in prior work~\cite{fu-etal-2024-qgeval} was insufficient to capture nuanced quality differences and resulted in lower inter annotator agreement.

\subsection{Human Evaluation}
We recruit 12 annotators in four language groups, with three native or fluent speakers per language. Annotators rate a stratified sample of 240 QA pairs (15 per model $\times$ 4 models $\times$ 4 languages), with three independent ratings per pair. The sample is drawn from 20 videos verified by human reviewers to contain high mentorship value, with five videos selected per language. Although this is a subset of the full dataset, the stratified design ensures equal representation across model and language conditions, enabling reliable comparisons. Larger scale annotation was not feasible because mentorship oriented QA requires native speaker judgment and domain familiarity. An example annotation interface is provided in \Cref{sec:eval}.

\paragraph{Inter Annotator Agreement.}
We measure agreement using Gwet's AC2, computed separately for each metric across languages and models. We use Gwet's AC2 rather than Krippendorff’s alpha because alpha can give misleading estimates under skewed rating distributions common in AI system evaluation~\cite{ebling-2024, pradhan-2025}.

\subsection{Automatic Evaluation}
Recent work shows that LLM based evaluators outperform traditional Natural Language Generation (NLG) metrics and provide a scalable alternative to human evaluation~\cite{fu-etal-2024-qgeval,kocmi-federmann-2023-large}. To study automatic evaluation for mentorship focused QA, we evaluate all QA pairs using nine multilingual LLM judges.

We select nine open source or publicly documented multilingual judges for reproducibility:
Qwen2.5 7B~\cite{yang2025qwen2.5-1m},
Command R 7B~\cite{cohere2025commandaenterprisereadylarge},
Mistral 8B Instruct~\cite{mistral8b_instruct_2410},
Llama3.2 11B~\cite{llama32_11b_vision},
Qwen3 8B~\cite{qwen3technicalreport},
Gemma3 12B IT~\cite{gemma3_12b_it},
Pangea 7B~\cite{yue2024pangeafullyopenmultilingual},
AyaExpanse 8B~\cite{dang2024ayaexpansecombiningresearch},
and Phi 4 14B~\cite{phi4_microsoft}. Only two of the nine judges belong to the Qwen family, while the remaining judges span several independent model families.

\paragraph{LLM Human Agreement.} 
We compute Gwet’s AC2 between LLM scores and human annotations for each metric, language, and QA generation model to quantify when automatic evaluation aligns with human judgments.

\section{Evaluation Results}
\label{sec:evaluation}
We evaluate mentorship focused QA generation along four axes: reliability of mentorship focused evaluation, scalability through LLM judges, effectiveness of QA generation models, and consistency across languages and topics.

\begin{figure}[t]
\centering
\includegraphics[width=0.87\columnwidth]{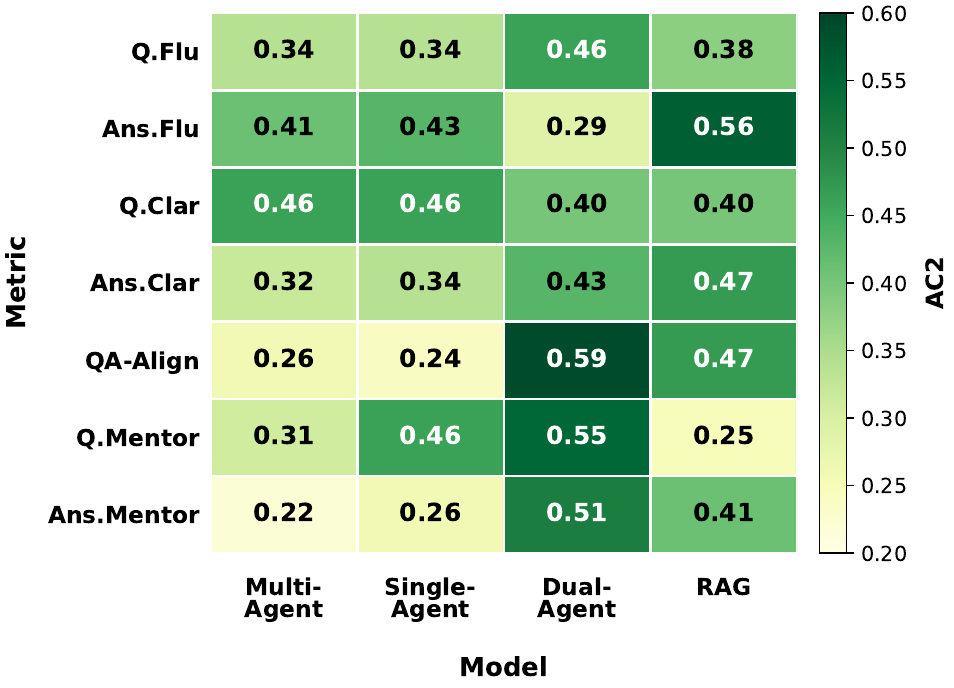}
\vspace{-0.8em}
\caption{IAA Scores.(Averaged across Languages)}
\label{fig:inter-ann-model}
\vspace{-0.5em}
\end{figure}

\paragraph{Reliability of Mentorship Focused Evaluation.}
Human annotations show the highest agreement for linguistic metrics: \textit{Fluency} (Q: $0.38$, A: $0.42$) and \textit{Clarity} (Q: $0.43$, A: $0.39$). Task oriented metrics also show stable moderate agreement: \textit{QA Alignment} ($0.39$) and \textit{Mentorship} (Q: $0.39$, A: $0.35$), reflecting the interpretive nature of mentorship quality. Agreement varies by model and language (Figure~\ref{fig:inter-ann-model}, language-averaged). \textit{Dual Agent shows the highest agreement} ($0.46$), followed by RAG ($0.42$), Single Agent ($0.36$), and Multi Agent ($0.33$). Across languages, agreement is highest for Romanian ($0.67$) and lower for Chinese ($0.37$), English ($0.28$), and Hindi ($0.26$) (more in \Cref{sec:eval}). These patterns show that mentorship focused evaluation is sensitive to model behavior and linguistic context, motivating culturally aware and mentorship specific evaluation.

\paragraph{Human LLM Agreement in Automatic Evaluation.}

\begin{figure}[t]
\centering
\includegraphics[width=1.0\columnwidth]{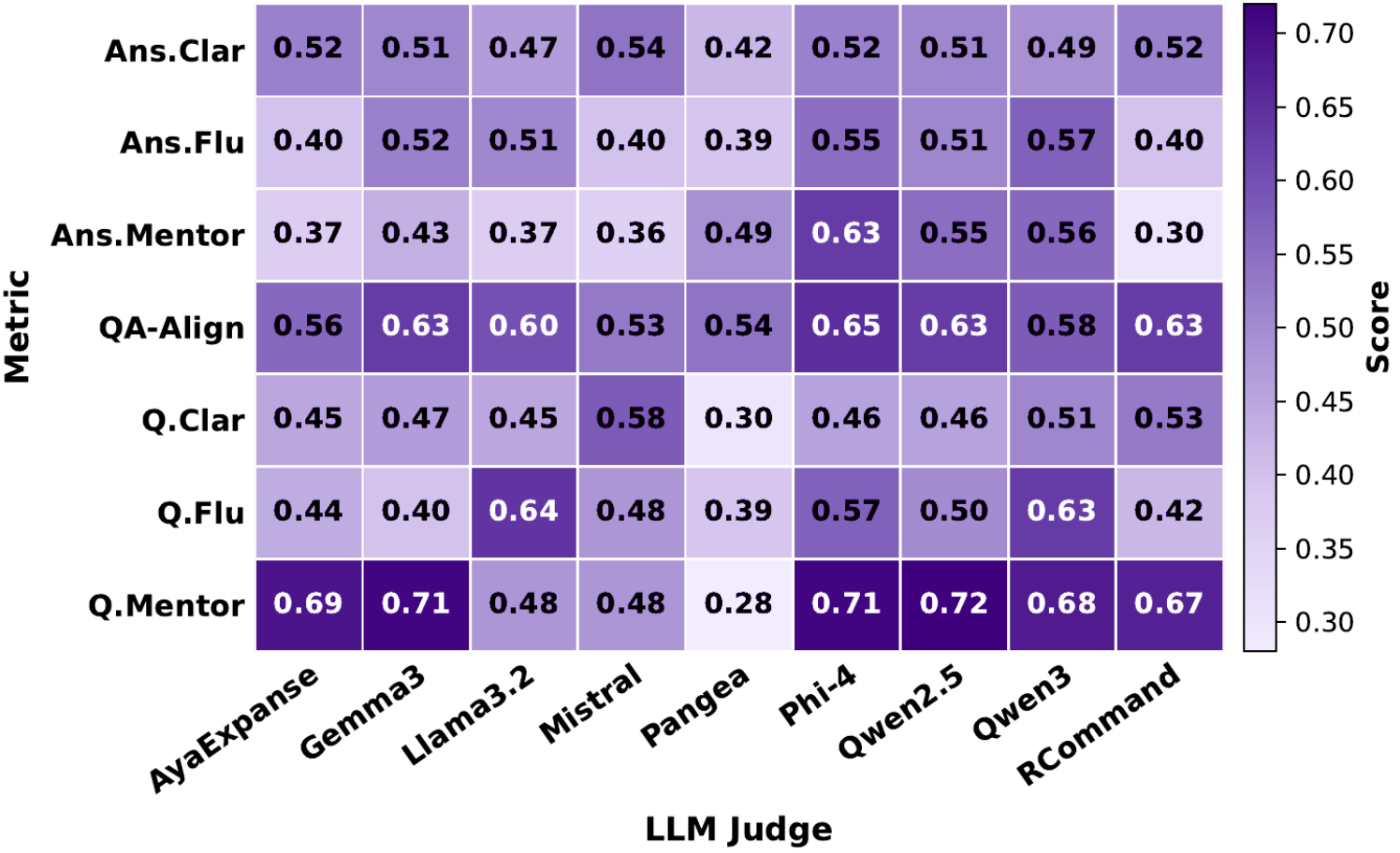}  
\caption{LLM-human agreement scores, with Qwen models, Phi-4, and Gemma3 aligning best.}
\label{fig:llm_human_agreement_avg}
\vspace{-1.5em}
\end{figure}

To assess scalability, we compare LLM based judges with human annotations in Figure~\ref{fig:llm_human_agreement_avg}. Agreement is highest for task oriented metrics, with Qwen2.5, Qwen3, Phi 4 and Gemma3 reaching substantial agreement with human ratings ($0.55$ to $0.72$), while linguistic metrics show moderate but stable alignment ($0.40$ to $0.57$). \textit{Qwen family models and Phi 4 achieve the strongest agreement across metrics and languages}, including lower resource settings (results per language in \Cref{sec:eval}). However, agreement varies substantially across judges, motivating language and metric aware evaluation with multiple LLM judges rather than a single evaluator. 

\paragraph{Transcript Quality.}
Since all pipelines use automatic transcripts, we assess transcript quality as an upstream factor. Whisper Large v3~\citep{radford2022robustspeechrecognitionlargescale} reports error rates of 9.3 for English, 10.8 for Romanian, 12.8 for Chinese, and 16.9 for Hindi. For Chinese, the reported value is computed at the character level, while the remaining languages use word level error rates. All four architectures receive identical transcripts, so transcription errors do not confound cross architecture comparisons. Annotators for English, Romanian, and Chinese did not flag transcription induced QA errors. Hindi annotators noted occasional issues consistent with its higher WER, but replacing Qwen2.5 7B with Gemma3 27B for Hindi (\Cref{sec:language_eval}) substantially improves agreement, suggesting base model capacity rather than transcript quality is the main bottleneck.

\paragraph{Comparison of QA Generation Architectures.}

\begin{figure}[h]
\centering
\includegraphics[width=0.82\columnwidth]{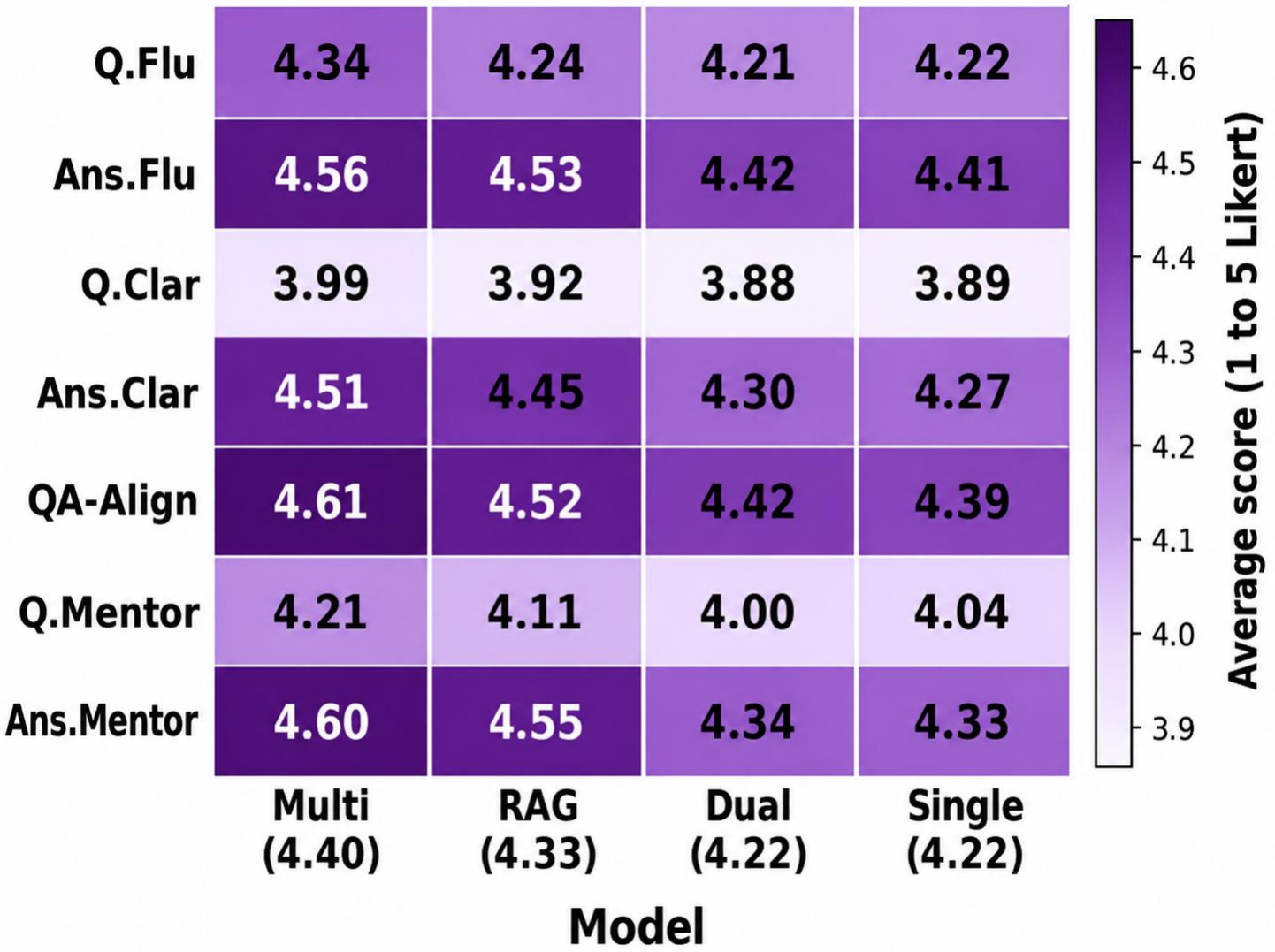}
\vspace{-0.8em}
\caption{QA generation performance across models and metrics (1 to 5 Likert). Models are sorted by mean.}
\label{fig:model_performance_heatmap}
\vspace{-0.5em}
\end{figure}
\vspace{-1em}
\begin{figure}[h]
\centering
\includegraphics[width=0.86\columnwidth]{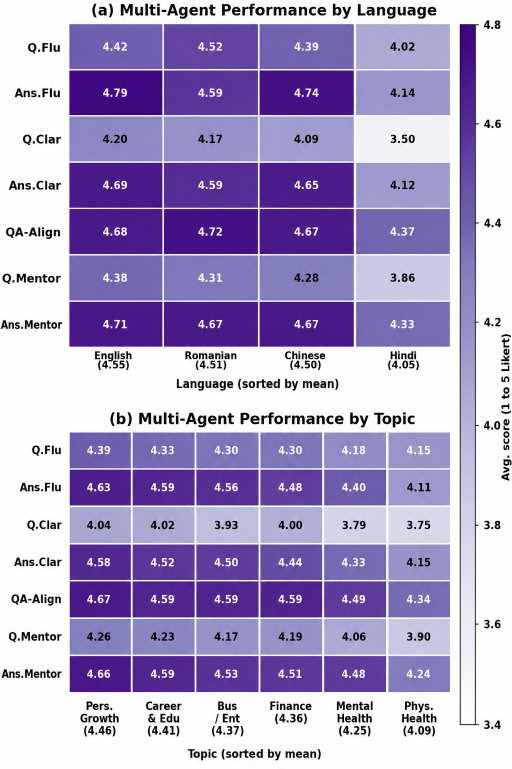}
\caption{Multi Agent performance across (a) language and (b) topic. Topics and languages are sorted by mean.}
\label{fig:multiagent_language_topic}
\vspace{-0.7em}
\end{figure}

The Multi Agent model performs best overall, ranking highest across all evaluation dimensions with a mean score of 4.40, as shown in \Cref{fig:model_performance_heatmap}. RAG follows closely at 4.33 and improves over Single Agent and Dual Agent, especially on \textit{QA Alignment}, \textit{Question Mentorship}, and \textit{Answer Mentorship}. Single Agent and Dual Agent perform nearly identically at 4.22, suggesting that topic aware chunking alone yields limited gains. Differences are largest for \textit{Mentorship} and \textit{QA Alignment}, while \textit{Fluency} remains similar across models. Detailed cross model and LLM judge results are in \Cref{sec:model-perf}.

\paragraph{Statistical Significance Analysis.}
We test architectural differences in human ratings using proportional odds Ordinal Logistic Regression (OLR) over the full 1 to 5 rating scale. This analysis preserves the ordering of all response categories rather than reducing the ratings to a single threshold. For the pooled analysis, the model includes language, annotator, and source video fixed effects. The language specific models include annotator and source video fixed effects. Standard errors are clustered by QA item because each QA pair receives three independent human ratings. We report odds ratios (ORs), where an OR greater than 1 indicates higher odds of receiving a better rating for Multi Agent relative to the comparison architecture. Holm correction is applied across the six planned Multi Agent contrasts for the two mentorship outcomes within each analysis. Table~\ref{tab:olr} reports the pooled results across all languages.

\begin{table}[ht]
\centering
\footnotesize
\setlength{\tabcolsep}{1.5pt}
\renewcommand{\arraystretch}{1.05}
\begin{tabular}{@{}lcc@{}}
\toprule
\textbf{Comparison} & \textbf{Q. Men.} & \textbf{A. Men.} \\
\midrule
Multi vs. Single &
1.88 [1.15, 3.08]$^{*}$ &
8.88 [5.66, 13.94]$^{**}$ \\

Multi vs. Dual &
4.38 [2.58, 7.44]$^{**}$ &
10.80 [6.38, 18.28]$^{**}$ \\

Multi vs. RAG &
1.57 [0.99, 2.47] &
1.74 [1.12, 2.71]$^{*}$ \\
\bottomrule
\end{tabular}

\caption{Pooled proportional odds OLR results across all languages. Cells report odds ratios with 95\% confidence intervals. Q. Men. denotes Question Mentorship and A. Men. denotes Answer Mentorship. Significance markers use Holm adjusted $p$ values across the six planned contrasts: $^{*}p<0.05$, $^{**}p<0.001$.}
\label{tab:olr}
\end{table}

Across all languages, the global Wald tests show significant architectural differences for both \textit{Question Mentorship} ($\chi^2(3)=30.33$, $p<0.001$) and \textit{Answer Mentorship} ($\chi^2(3)=123.26$, $p<0.001$).

For \textit{Question Mentorship}, Multi Agent receives significantly higher ratings than Single Agent (OR$=1.88$, 95\% CI $[1.15,3.08]$, Holm adjusted $p=0.037$) and Dual Agent (OR$=4.38$, 95\% CI $[2.58,7.44]$, Holm adjusted $p<0.001$). The difference between Multi Agent and RAG is smaller and does not remain significant after correction (OR$=1.57$, 95\% CI $[0.99,2.47]$, Holm adjusted $p=0.055$).

For \textit{Answer Mentorship}, Multi Agent has substantially higher odds of receiving better ratings than Single Agent (OR$=8.88$, 95\% CI $[5.66,13.94]$, Holm adjusted $p<0.001$) and Dual Agent (OR$=10.80$, 95\% CI $[6.38,18.28]$, Holm adjusted $p<0.001$). Multi Agent also significantly outperforms RAG in the pooled analysis (OR$=1.74$, 95\% CI $[1.12,2.71]$, Holm adjusted $p=0.037$). These results provide strong ordinal evidence that the Multi Agent architecture improves the mentorship quality of generated answers.

Because the pooled effect is particularly strong for \textit{Answer Mentorship}, we further examine whether this pattern is observed across individual languages. The global architecture effect is significant for Answer Mentorship in English ($\chi^2(3)=40.15$, $p<0.001$), Chinese ($\chi^2(3)=22.25$, $p<0.001$), Hindi ($\chi^2(3)=26.00$, $p<0.001$), and Romanian ($\chi^2(3)=40.98$, $p<0.001$). As shown in Table~\ref{tab:olr_language_answer_a}(a) and Table~\ref{tab:olr_language_answer_b}(b), Multi Agent receives significantly higher Answer Mentorship ratings than both Single Agent and Dual Agent in every language after Holm correction. The contrasts with RAG are smaller and do not reach significance after correction within the individual language analyses.

\begin{table}[ht]
\centering
\footnotesize
\setlength{\tabcolsep}{4pt}
\renewcommand{\arraystretch}{1.08}
\begin{tabular}{lccc}
\toprule
\textbf{Comparison} & \textbf{OR} & \textbf{95\% CI} & \textbf{Holm $p$} \\
\midrule
\multicolumn{4}{l}{\textbf{English}} \\
Multi Agent vs. Single & 7.68 & [3.57, 16.50] & $<0.001$ \\
Multi Agent vs. Dual   & 45.46 & [12.63, 163.62] & $<0.001$ \\
Multi Agent vs. RAG    & 2.17 & [0.91, 5.14] & 0.158 \\
\midrule
\multicolumn{4}{l}{\textbf{Chinese}} \\
Multi Agent vs. Single & 3.53 & [1.55, 8.07] & 0.011 \\
Multi Agent vs. Dual   & 6.16 & [2.44, 15.54] & $<0.001$ \\
Multi Agent vs. RAG    & 0.96 & [0.36, 2.53] & 0.932 \\
\bottomrule
\end{tabular}
\caption{(a) Language specific proportional odds OLR results for Answer Mentorship in English and Chinese. Each comparison reports Multi Agent relative to the indicated architecture. Holm adjusted $p$ values are computed across the six planned mentorship contrasts within each language.}
\label{tab:olr_language_answer_a}
\end{table}

\begin{table}[ht]
\centering
\footnotesize
\setlength{\tabcolsep}{4pt}
\renewcommand{\arraystretch}{1.08}
\begin{tabular}{lccc}
\toprule
\textbf{Comparison} & \textbf{OR} & \textbf{95\% CI} & \textbf{Holm $p$} \\
\midrule
\multicolumn{4}{l}{\textbf{Hindi}} \\
Multi Agent vs. Single & 7.19 & [2.74, 18.89] & $<0.001$ \\
Multi Agent vs. Dual   & 6.14 & [2.74, 13.79] & $<0.001$ \\
Multi Agent vs. RAG    & 2.59 & [1.18, 5.71] & 0.072 \\
\midrule
\multicolumn{4}{l}{\textbf{Romanian}} \\
Multi Agent vs. Single & 63.81 & [17.79, 228.93] & $<0.001$ \\
Multi Agent vs. Dual   & 17.81 & [3.75, 84.55] & 0.001 \\
Multi Agent vs. RAG    & 1.74 & [0.65, 4.68] & 0.280 \\
\bottomrule
\end{tabular}

\addtocounter{table}{-1}
\caption{(b) Language specific proportional odds OLR results for Answer Mentorship in Hindi and Romanian. Each comparison reports Multi Agent relative to the indicated architecture. Holm adjusted $p$ values are computed across the six planned mentorship contrasts within each language.}
\label{tab:olr_language_answer_b}
\end{table}

The language specific analysis therefore reinforces the pooled result for Answer Mentorship. Although the estimated effect size varies across languages, Multi Agent shows a consistent and statistically significant advantage over Single Agent and Dual Agent. Importantly, these results are based entirely on human ratings, providing independent support for the mentorship gains observed in the broader evaluation and showing that the effect cannot be explained solely by a preference for longer responses among LLM judges.

\paragraph{Qualitative Analysis.} 
Tables~\ref{tab:qualitative_english}, \ref{tab:qualitative_chinese}, \ref{tab:qualitative_hindi}, and \ref{tab:qualitative_romanian} show representative QA pairs across the four languages. Single Agent responses are often brief and factual, while Multi Agent responses provide more structured, point-wise actionable guidance, which human annotators rated as more helpful across languages.

\begin{table}[ht]
\centering
\scriptsize
\setlength{\tabcolsep}{3pt}
\renewcommand{\arraystretch}{1.3}
\begin{tabular}{p{0.8cm} p{5.0cm} p{0.6cm} p{0.6cm}}
\toprule
\textbf{Model} & \textbf{Question / Answer} & \textbf{Q.Men} & \textbf{A.Men} \\
\midrule
Single
& \textbf{Q:} Can generative models be used for creating content in low-resource languages? \newline
  \textbf{A:} Generative models do not work well for low-resource languages, generating only garbage... should be used cautiously by native speakers only.
& 4 & 2 \\
\midrule
Dual
& \textbf{Q:} What are some areas of research that the speaker admires? \newline
  \textbf{A:} The speaker admires research in Guarani papers... and OCR for low-resource and parsing papers.
& 3 & 3 \\
\midrule
RAG
& \textbf{Q:} Can you provide examples of successful low-resource language projects? \newline
  \textbf{A:} Examples include text-to-speech systems... for Tunisian Arabic and Korean... surprisingly low-resource despite more digital resources.
& 4 & 3 \\
\midrule
Multi
& \textbf{Q:} What strategies overcome challenges collaborating with native speakers of low-resource languages? \newline
  \textbf{A:} (1)~\textit{Patience}: reaching out requires time and effort... (2)~\textit{Building Trust}: show genuine interest and respect... (3)~\textit{Community Protocols}: understand each community's norms... (4)~\textit{Tangible Benefits}: develop tools that directly help the community.
& 5 & 5 \\
\bottomrule
\end{tabular}
\caption{English QA examples across architectures.}
\label{tab:qualitative_english}
\end{table}

\begin{table}[ht]
\centering
\scriptsize
\setlength{\tabcolsep}{3pt}
\renewcommand{\arraystretch}{1.3}
\begin{tabular}{p{0.8cm} p{5.0cm} p{0.6cm} p{0.6cm}}
\toprule
\textbf{Model} & \textbf{Question / Answer} & \textbf{Q.Men} & \textbf{A.Men} \\
\midrule
Single
& \textbf{Q:} 更年期的定义是什么？ \newline
  \textbf{A:} 更年期就是月经周期的结束，即你最后一次月经。
& 3 & 2 \\
\midrule
Dual
& \textbf{Q:} 什么是脑雾（Brain fog）？ \newline
  \textbf{A:} 脑雾是指大脑似乎不属于你... 忘记想说的话... 注意力不集中，感觉大脑里有雾。
& 3 & 3 \\
\midrule
RAG
& \textbf{Q:} 为什么更年期被认为是人生中的重要阶段？ \newline
  \textbf{A:} 更年期标志着最后一次月经... 意味着生育能力的结束... 对身体和心理都会产生显著影响，包括情绪波动、记忆力减退等。
& 4 & 4 \\
\midrule
Multi
& \textbf{Q:} 如何从三十几岁起为应对更年期做好心理和生理上的准备？ \newline
  \textbf{A:} (1)~\textit{了解更年期}：认识到这是女性生命中的自然阶段... (2)~\textit{健康生活方式}：保持健康饮食，定期锻炼... (3)~\textit{管理压力}：通过冥想、瑜伽应对情绪波动... (4)~\textit{建立支持系统}：与家人朋友分享感受。
& 4 & 5 \\
\bottomrule
\end{tabular}
\caption{Chinese QA examples across architectures.}
\label{tab:qualitative_chinese}
\vspace{-5pt}
\end{table}

\begin{table}[ht]
\centering
\scriptsize
\setlength{\tabcolsep}{3pt}
\renewcommand{\arraystretch}{1.3}
\begin{tabular}{p{0.8cm} p{5.0cm} p{0.6cm} p{0.6cm}}
\toprule
\textbf{Model} & \textbf{Question / Answer} & \textbf{Q.Men} & \textbf{A.Men} \\
\midrule
Single
& \textbf{Q:} \begin{hindi}आपके चार्टर्ड अकाउंटेंट की यात्रा की शुरुआत कैसी थी?\end{hindi} \newline
  \textbf{A:} \begin{hindi}आपकी यात्रा की शुरुआत बहुत दिलचस्प थी।\end{hindi} [degrades into incoherent repetition]
& 3 & 1 \\
\midrule
Dual
& \textbf{Q:} \begin{hindi}चार्टर्ड अकाउंटेंट की यात्रा के बारे में आपका उपाय क्या था?\end{hindi} \newline
  \textbf{A:} \begin{hindi}पहले विद्यालय में अच्छा छात्र रहा... फाउंडेशन के लिए एक निश्चित स्ट्रेटेजी अपनाई थी।\end{hindi}
& 3 & 2 \\
\midrule
RAG
& \textbf{Q:} \begin{hindi}फाउंडेशन एक्जाम के लिए किस प्रकार की प्रैक्टिस की थी?\end{hindi} \newline
  \textbf{A:} \begin{hindi}साइमल्टेनियस प्रैक्टिस किया... थियोरिकल सब्जेक्ट्स से शुरू... अकाउंटिंग स्टैंडर्ड्स और प्रोफेसर के नोट्स का उपयोग किया।\end{hindi}
& 3 & 3 \\
\midrule
Multi
& \textbf{Q:} \begin{hindi}परीक्षा खतम होने के बाद परिणाम की तैयारी के लिए समय कैसे उपयोग करें?\end{hindi} \newline
  \textbf{A:} \begin{hindi}एक नियमित प्रोग्राम बनाएं। थोड़ा आराम लें, फिर विषय-विषयी सारणी बनाएं। पिछली परीक्षाओं की जांच करें और त्रुटियों को ठीक करें। बुक्स और वीडियो का उपयोग करें।\end{hindi}
& 4 & 5 \\
\bottomrule
\end{tabular}
\caption{Hindi QA examples across architectures.}
\label{tab:qualitative_hindi}
\end{table}

\begin{table}[ht]
\centering
\scriptsize
\setlength{\tabcolsep}{3pt}
\renewcommand{\arraystretch}{1.3}
\begin{tabular}{p{0.8cm} p{5.0cm} p{0.6cm} p{0.6cm}}
\toprule
\textbf{Model} & \textbf{Question / Answer} & \textbf{Q.Men} & \textbf{A.Men} \\
\midrule
Single
& \textbf{Q:} Ce a fost un moment în care te simțeai backstabbed? \newline
  \textbf{A:} Un moment a fost când am făcut documentarea... și am reușit să facem o bună documentare.
& 3 & 2 \\
\midrule
Dual
& \textbf{Q:} Cum a influențat popularitatea seriei viața ta înainte de 20 de ani? \newline
  \textbf{A:} Popularitatea m-a dus la notorietate... viața era localizată înainte... cu succesul seriei m-au cunoscut mai mulți oameni.
& 3 & 3 \\
\midrule
RAG
& \textbf{Q:} Cum te simți când oferi timp lucrurilor care te aduc bucurie? \newline
  \textbf{A:} Mă simt realizat și plin de energie... bucuria vine din conexiunea cu oamenii... un sentiment de pură bucurie în fiecare zi.
& 3 & 4 \\
\midrule
Multi
& \textbf{Q:} Cum ai merge în pregătirea pentru a face un curs de actorie? \newline
  \textbf{A:} Este important să te implici profund... să nu fii satisfăcut de rezultate imediate... să te străduiești pentru calitate... dedicația și lipsa de superficialitate sunt esențiale.
& 4 & 5 \\
\bottomrule
\end{tabular}
\caption{Romanian QA examples across architectures.}
\label{tab:qualitative_romanian}
\end{table}

\paragraph{Language Evaluation.}\label{sec:language_eval}
Using the best performing Multi Agent model with scores averaged over all LLM judges, we analyze language effects in \Cref{fig:multiagent_language_topic}(a), with cross model comparisons in \Cref{sec:eval}. \textit{English performs best overall (4.55)}, while \textit{Romanian (4.51) and Chinese (4.50)} remain comparable on \textit{QA Alignment} and \textit{Answer Mentorship}. \textit{Hindi scores lower overall (4.05)}, especially on \textit{Question Clarity} (3.50) and \textit{Question Mentorship   } (3.86), consistent with Devanagari tokenization and training limitations, as well as MMLU results where Hindi trails English~\cite{qwen25_14b_hindi}. However, Multi Agent narrows this gap: compared with Single Agent scores of 3.96 on QA Alignment and 3.72 on Answer Fluency, it improves Hindi to 4.37 and 4.14, suggesting that agentic coordination can partially mitigate language specific errors.

\begin{figure}[!t]
\centering
\vspace{-0.2em}
\includegraphics[width=0.95\columnwidth]{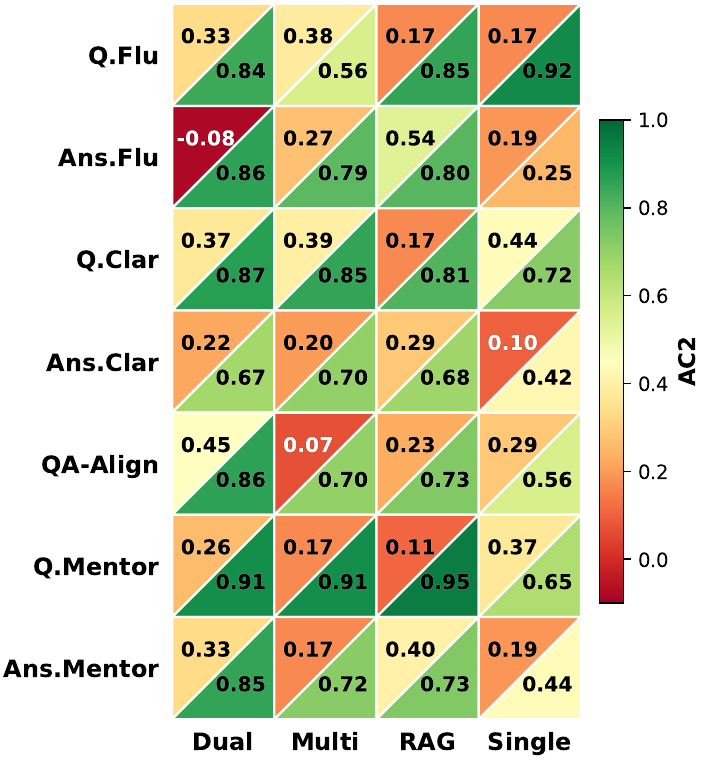}
\vspace{-0.2em}
\caption{Hindi IAA: Qwen2.5 7B (upper \upperhalftri) vs.\ Gemma3 27B (lower \lowerhalftri) across models and metrics.}
\label{fig:hindi_iaa}
\vspace{-1.5em}
\end{figure}

To validate that low Hindi IAA reflects base model limitations rather than flaws in our evaluation framework, we replace Qwen2.5 7B with Gemma3 27B. As shown in \Cref{fig:hindi_iaa}, IAA rises from near chance agreement to substantial agreement (0.56 to 0.95). Multi Agent consistently outperforms Single Agent under both models, confirming that agentic coordination provides genuine quality gains independent of model size.

\paragraph{Topic Evaluation.}
Using the same Multi Agent setting, we analyze topic effects in \Cref{fig:multiagent_language_topic}(b). \textit{Personal Growth} performs best overall (4.46), with high \textit{QA Alignment} (4.67) and \textit{Answer Mentorship} (4.66). \textit{Career \& Education} (4.41) and \textit{Business/Enterprise} (4.37) follow, while \textit{Physical Health} scores lowest (4.09), especially on \textit{Question Clarity} (3.75) and \textit{Question Mentorship} (3.90), reflecting clinically dense language. \textit{Finance} remains strong on \textit{QA Alignment} (4.47), likely due to structured discourse. Overall, agentic coordination appears most useful for complex mentorship topics, with smaller gains for domains well represented in pretraining.

\paragraph{Takeaways.}
Our results show that: (1) task oriented metrics are reliable but more subjective than linguistic metrics; (2) LLM evaluation must be language and metric aware, motivating multi judge strategies; (3) Multi Agent models outperform simpler architectures on mentorship and alignment; and (4) agentic coordination is especially useful for complex topics and lower resource languages.

\section{Conclusion}

We introduced \textsc{MentorQA}, the first multilingual dataset and evaluation framework for mentorship focused question answering from long form videos. By treating mentorship as a distinct QA objective, we show that existing QA benchmarks remain limited for settings that require guidance, clarity, and learning value beyond factual correctness.

Through a controlled comparison of Single Agent, Dual Agent, RAG, and Multi Agent pipelines, we find that agentic coordination consistently improves mentorship quality, especially for complex topics and lower resource languages. Building on these results, a promising future direction is a hybrid architecture that combines RAG's efficient retrieval and chunking with Multi Agent question scoring, ranking, and brainstorming, preserving quality gains while reducing cost.

Our automated evaluation analysis further shows substantial variation in alignment with human judgments across languages and metrics, highlighting the need for task aware and language sensitive evaluation. Overall, \textsc{MentorQA} establishes mentorship focused QA as a new research direction and provides a multilingual benchmark for studying architectures and evaluation methods that better support learning, reflection, and guidance across cultures.
The dataset and models are released at \url{https://aim-scu.github.io/MentorQA}.

\section*{Limitations}
Our work represents an initial step toward mentorship-focused question answering. While it provides a comprehensive evaluation of QA architectures, several limitations remain.

\noindent\textbf{Generality of Findings.}
Our study focuses on mentorship-focused QA derived from long-form videos across four languages and a limited set of domains. This design enables controlled comparison of architectures and evaluation dimensions, but it does not capture the full diversity of mentorship contexts, interaction styles, or cultural norms found in real-world settings. In particular, our findings may not transfer to highly interactive mentorship scenarios or to languages with substantially different resource profiles. Rather than aiming for universal coverage, our goal is to establish mentorship-focused QA as a distinct problem and to provide a benchmark that supports further evaluation and analysis. We encourage future work to extend our open-source framework to additional languages and domains.

\noindent\textbf{A Fixed Base Model.}
To isolate the effects of architectural and agentic design, we fix the underlying base language model (Qwen~2.5) across all experiments. This choice enables fair comparison but limits insight into how multi-agent pipelines interact with other foundation models. Newer multilingual or instruction-tuned models may exhibit different trade-offs or amplify agentic gains, which we invite future work to explore.

\noindent\textbf{Allocation Design.}
To enable controlled comparison across architectures, we use a fixed question budget of $T=20$ and a single segment strength function, $S(i)=\mu-\sigma$, throughout the Multi Agent pipeline. These choices keep the generation setting consistent, but we do not explore alternative allocation functions or question budgets in this work. Future work could examine whether different allocation strategies or values of $T$ further improve mentorship quality or coverage.

\noindent\textbf{Evaluation and Scalability.}
Our evaluation combines human judgments with LLM-based judges to assess mentorship quality across multiple dimensions. Although LLM judges provide scalable signal, we observe notable variation in their alignment with human judgments across languages and metrics. Accordingly, automated evaluation should be viewed as complementary rather than a substitute for human assessment. Multi-agent pipelines also incur higher computational cost than single-agent or RAG-based approaches, which may limit deployment in resource-constrained settings. Our analysis prioritizes comparative effectiveness, and we see efficiency-aware agentic designs as an important direction for future work.

\noindent\textbf{Subjectivity of Mentorship Evaluation.}
Mentorship quality is inherently subjective, as reflected in the moderate inter-annotator agreement observed for mentorship-focused metrics. Annotators may reasonably prioritize different aspects of guidance, particularly across cultural and linguistic contexts. While our evaluation captures this variability, it does not explicitly model disagreement or multiple valid interpretations of mentorship quality. Developing evaluation protocols that better account for such diversity remains an open challenge.

\noindent\textbf{Downstream Educational Impact.}
Our evaluation focuses on mentorship QA quality rather than downstream educational outcomes. Future work could examine how mentorship-focused QA supports comprehension, reflection, and skill development in real educational settings.


\noindent\textbf{Computational Cost.}
On average per 1.5-hour video, Single-Agent requires 78 seconds, Dual-Agent 130 seconds, RAG 181 seconds, and Multi-Agent 316 seconds. The increased latency in Multi-Agent is primarily driven by the Question Brainstorming and Scoring steps, which involve over-generation and iterative evaluation. This overhead is justified by the substantial mentorship quality gains, particularly for lower-resource languages and complex topics where simpler architectures degrade significantly, making it a worthwhile trade-off for educational AI applications. As discussed in the Conclusion, a hybrid architecture combining RAG efficiency with Multi-Agent coordination remains a promising direction to reduce this cost.

\section*{Ethical Considerations}
Mentorship content often involves personal experiences, sensitive advice, and implicit power dynamics. Although all videos used in \textsc{MentorQA} are publicly available, we emphasize that mentorship quality is not equivalent to factual authority, and generated QA pairs should not be treated as prescriptive guidance without human oversight. Automated systems may overgeneralize advice, miss contextual nuance, or reflect cultural biases present in training data.

We also note ethical risks in automated evaluation. LLM-based judges may encode language- or culture-specific preferences, which can disadvantage certain communication styles or communities. Our findings underscore the continued importance of human evaluation, transparent metric design, and cautious interpretation of automatic scores, particularly in multilingual and educational settings.

We release \textsc{MentorQA} as a research resource rather than a deployable mentorship system, and encourage future work to incorporate participatory evaluation, educator-in-the-loop validation, and safeguards that prioritize learner well-being.

\paragraph{Anonymization and Personal Information}
\label{sec:pii}
\textsc{MentorQA} is constructed from publicly available mentorship videos.
While the source content is public, we take steps to minimize the presence and propagation of personally identifiable information (PII) in the released dataset.
We process only automatically generated transcripts and do not release raw audio or video.
The released dataset contains only de-identified text and focuses on general mentorship themes, guidance, and reflective insights rather than personal narratives tied to specific individuals.
We follow established best practices for ethical data use from publicly available sources and release the dataset for research purposes only.

\section*{Acknowledgements}
We thank the anonymous reviewers for their constructive feedback. We also thank the members of the Inclusive and Multimodal AI (AIM) Lab at Santa Clara University for their help during the annotation process. This project was partially funded by an HPC Center Catalytic Grant and a Fletcher Jones Foundation grant.

\bibliography{custom}

\appendix

\section{Data Collection}
\label{sec:data}

\subsection{Video Sources}
\label{sec:videos}
To see the full detailed source of all videos, see Table \ref{tab:en_ch_videos} and Table \ref{tab:hi_ro_videos}. 

\begin{table*}[t]
\centering
\caption{Mentorship Videos}
\label{tab:en_ch_videos}
\begin{tabular}{|c c c|c c c|}
\hline
\textbf{Video ID} & \textbf{URL} & \textbf{Language} & \textbf{Video ID} & \textbf{URL} & \textbf{Language} \\
\hline

1  & \href{https://www.youtube.com/watch?v=jFWF0PX4Pcs}{Video Link} & English & 31 & \href{https://youtu.be/uFmKEbMgTzY}{Video Link} & Chinese \\
2  & \href{https://www.youtube.com/watch?v=pOJuyl4xOEg}{Video Link} & English & 32 & \href{https://youtu.be/fEyy-mNnWak}{Video Link} & Chinese \\
3  & \href{https://www.youtube.com/watch?v=TwQD6YeSXPQ}{Video Link} & English & 33 & \href{https://youtu.be/H6S2gM0mIpg}{Video Link} & Chinese \\
4  & \href{https://www.youtube.com/watch?v=nyHwi7x7t20}{Video Link} & English & 34 & \href{https://youtu.be/T1jhxIlHN-U}{Video Link} & Chinese \\
5  & \href{https://www.youtube.com/watch?v=40W6hoonzKg}{Video Link} & English & 35 & \href{https://youtu.be/LDJh_1tp9OE}{Video Link} & Chinese \\
6  & \href{https://www.youtube.com/watch?v=qNiBfZYWMH4}{Video Link} & English & 36 & \href{https://youtu.be/4YNLDWbj3OI}{Video Link} & Chinese \\
7  & \href{https://www.youtube.com/watch?v=ML5wUHyjSuU}{Video Link} & English & 37 & \href{https://youtu.be/omDnvoV066g}{Video Link} & Chinese \\
8  & \href{https://www.youtube.com/watch?v=kdZZHKNn3PQ}{Video Link} & English & 38 & \href{https://youtu.be/pDxi9MljStw}{Video Link} & Chinese \\
9  & \href{https://www.youtube.com/watch?v=u7kCuH2F8i4}{Video Link} & English & 39 & \href{https://youtu.be/s3Ciz7K9Ir8}{Video Link} & Chinese \\
10 & \href{https://www.youtube.com/watch?v=xk4TyzJqitU}{Video Link} & English & 40 & \href{https://youtu.be/KJbDwHnWq0s}{Video Link} & Chinese \\
11 & \href{https://www.youtube.com/watch?v=CFQz03C2Pq8}{Video Link} & English & 41 & \href{https://youtu.be/JzASHmY0oww}{Video Link} & Chinese \\
12 & \href{https://www.youtube.com/watch?v=ze7Zw0VOgGI}{Video Link} & English & 42 & \href{https://youtu.be/7hOQe3A_YMQ}{Video Link} & Chinese \\
13 & \href{https://www.youtube.com/watch?v=E1S6ep00aE4}{Video Link} & English & 43 & \href{https://youtu.be/KZQ3cbJpj54}{Video Link} & Chinese \\
14 & \href{https://www.youtube.com/watch?v=gRXxe2OxZV0}{Video Link} & English & 44 & \href{https://youtu.be/qtXsWBZgnN8}{Video Link} & Chinese \\
15 & \href{https://www.youtube.com/watch?v=tvN7LHdc-mw}{Video Link} & English & 45 & \href{https://youtu.be/c12QWJ7qVC0}{Video Link} & Chinese \\
16 & \href{https://www.youtube.com/watch?v=oBFYH6Ko_nE}{Video Link} & English & 46 & \href{https://youtu.be/2aYCUfRXPXk}{Video Link} & Chinese \\
17 & \href{https://www.youtube.com/watch?v=meMKVeUdvY8}{Video Link} & English & 47 & \href{https://youtu.be/pHBu10k1RKQ}{Video Link} & Chinese \\
18 & \href{https://www.youtube.com/watch?v=u5RVVChloIM}{Video Link} & English & 48 & \href{https://youtu.be/P66sAojAWwI}{Video Link} & Chinese \\
19 & \href{https://www.youtube.com/watch?v=g59P3YiqRxU}{Video Link} & English & 49 & \href{https://youtu.be/CxEcjK5StHs}{Video Link} & Chinese \\
20 & \href{https://www.youtube.com/watch?v=TlexQmpJtAE}{Video Link} & English & 50 & \href{https://youtu.be/MPpXf4CYZDA}{Video Link} & Chinese \\
21 & \href{https://www.youtube.com/watch?v=WQF_m4kD32w}{Video Link} & English & 51 & \href{https://youtu.be/8JHlW7aTkdU}{Video Link} & Chinese \\
22 & \href{https://www.youtube.com/watch?v=lsoGzWh0HUw}{Video Link} & English & 52 & \href{https://youtu.be/qyOvO5x8UcQ}{Video Link} & Chinese \\
23 & \href{https://www.youtube.com/watch?v=s0b3OTAg_h0}{Video Link} & English & 53 & \href{https://youtu.be/LEVd0jaJ_a0}{Video Link} & Chinese \\
24 & \href{https://www.youtube.com/watch?v=NbVZeAf155w}{Video Link} & English & 54 & \href{https://youtu.be/EoyaiwVEw4k}{Video Link} & Chinese \\
25 & \href{https://www.youtube.com/watch?v=5VZyscu57_U}{Video Link} & English & 55 & \href{https://youtu.be/VI4rSYYDOQ4}{Video Link} & Chinese \\
26 & \href{https://www.youtube.com/watch?v=A1KTMMfNzdA}{Video Link} & English & 56 & \href{https://youtu.be/NX8dz2YPNaY}{Video Link} & Chinese \\
27 & \href{https://www.youtube.com/watch?v=3dzYvDLHbbs}{Video Link} & English & 57 & \href{https://youtu.be/tqa2kNicuK4}{Video Link} & Chinese \\
28 & \href{https://www.youtube.com/watch?v=6v6Ba4DQcMI}{Video Link} & English & 58 & \href{https://youtu.be/LdSHaX9SSVo}{Video Link} & Chinese \\
29 & \href{https://www.youtube.com/watch?v=qCYR9seVkEM}{Video Link} & English & 59 & \href{https://youtu.be/LqOlkE2Y0bQ}{Video Link} & Chinese \\
30 & \href{https://www.youtube.com/watch?v=rl8ZhQgyK3I}{Video Link} & English & 60 & \href{https://youtu.be/RoVh6aQ_TgA}{Video Link} & Chinese \\
\hline
\end{tabular}
\end{table*}

\begin{table*}[t]
\centering
\caption{Mentorship Videos (Continued)}
\label{tab:hi_ro_videos}
\begin{tabular}{|c c c|c c c|}
\hline
\textbf{Video ID} & \textbf{URL} & \textbf{Language} & \textbf{Video ID} & \textbf{URL} & \textbf{Language} \\
\hline

61 & \href{https://www.youtube.com/watch?v=vb0cPkDegpw}{Video Link} & Hindi & 91  & \href{https://www.youtube.com/watch?v=UvI5mOqOEhQ}{Video Link} & Romanian \\
62 & \href{https://www.youtube.com/watch?v=EuMR4pdO4PU}{Video Link} & Hindi & 92  & \href{https://www.youtube.com/watch?v=qUH1E1YqmFo}{Video Link} & Romanian \\
63 & \href{https://www.youtube.com/watch?v=Yvx624zKSB8}{Video Link} & Hindi & 93  & \href{https://www.youtube.com/watch?v=LO_W5aJY58Y}{Video Link} & Romanian \\
64 & \href{https://www.youtube.com/watch?v=YG1sW00jwLY}{Video Link} & Hindi & 94  & \href{https://www.youtube.com/watch?v=BeSUMgcW6vM}{Video Link} & Romanian \\
65 & \href{https://www.youtube.com/watch?v=7KxjHxj3rKk}{Video Link} & Hindi & 95  & \href{https://www.youtube.com/watch?v=eDBpZfwPB9Q}{Video Link} & Romanian \\
66 & \href{https://www.youtube.com/watch?v=jccPTbquRLk}{Video Link} & Hindi & 96  & \href{https://www.youtube.com/watch?v=XDpgET9XBFc}{Video Link} & Romanian \\
67 & \href{https://www.youtube.com/watch?v=5ZYEq6-5RwI}{Video Link} & Hindi & 97  & \href{https://www.youtube.com/watch?v=EubOx_UuOs8}{Video Link} & Romanian \\
68 & \href{https://www.youtube.com/watch?v=3CntDXz7jok}{Video Link} & Hindi & 98  & \href{https://www.youtube.com/watch?v=zAqx6OMyt28}{Video Link} & Romanian \\
69 & \href{https://www.youtube.com/watch?v=fHBR1j1kJ1I}{Video Link} & Hindi & 99  & \href{https://www.youtube.com/watch?v=HqiikIj1rWQ}{Video Link} & Romanian \\
70 & \href{https://www.youtube.com/watch?v=hsMN1Pi9Rbs}{Video Link} & Hindi & 100 & \href{https://www.youtube.com/watch?v=1iIzlcOYgEY}{Video Link} & Romanian \\
71 & \href{https://www.youtube.com/watch?v=MFyEwdpC5pM}{Video Link} & Hindi & 101 & \href{https://www.youtube.com/watch?v=EVTGtfT4HhE}{Video Link} & Romanian \\
72 & \href{https://www.youtube.com/watch?v=pKnFNwGL2Tw}{Video Link} & Hindi & 102 & \href{https://www.youtube.com/watch?v=X9Q7ckeEhIw}{Video Link} & Romanian \\
73 & \href{https://www.youtube.com/watch?v=sMuheiO3E-U}{Video Link} & Hindi & 103 & \href{https://www.youtube.com/watch?v=jT3TjuLZ7lE}{Video Link} & Romanian \\
74 & \href{https://www.youtube.com/watch?v=D74vLgMYOxM}{Video Link} & Hindi & 104 & \href{https://www.youtube.com/watch?v=frnTIOfMrCk}{Video Link} & Romanian \\
75 & \href{https://www.youtube.com/watch?v=kb2cEb20KXE}{Video Link} & Hindi & 105 & \href{https://www.youtube.com/watch?v=jmm9FOEpjwk}{Video Link} & Romanian \\
76 & \href{https://www.youtube.com/watch?v=210uLW5dh8w}{Video Link} & Hindi & 106 & \href{https://www.youtube.com/watch?v=sID4SFc_vR8}{Video Link} & Romanian \\
77 & \href{https://www.youtube.com/watch?v=S_0ZP51eIkE}{Video Link} & Hindi & 107 & \href{https://www.youtube.com/watch?v=_UnkDKXGavQ}{Video Link} & Romanian \\
78 & \href{https://www.youtube.com/watch?v=4xg2FOscsVE}{Video Link} & Hindi & 108 & \href{https://www.youtube.com/watch?v=dZXph1RHT2I}{Video Link} & Romanian \\
79 & \href{https://www.youtube.com/watch?v=hTJUBvPiviM}{Video Link} & Hindi & 109 & \href{https://www.youtube.com/watch?v=ckns_7K5p_s}{Video Link} & Romanian \\
80 & \href{https://www.youtube.com/watch?v=Z9Pam0MDl-4}{Video Link} & Hindi & 110 & \href{https://www.youtube.com/watch?v=pToJQ0RjpJY}{Video Link} & Romanian \\
81 & \href{https://www.youtube.com/watch?v=ADm9p803kWM}{Video Link} & Hindi & 111 & \href{https://www.youtube.com/watch?v=CAMbWv1vpj0}{Video Link} & Romanian \\
82 & \href{https://www.youtube.com/watch?v=MpDe3KDhYUg}{Video Link} & Hindi & 112 & \href{https://www.youtube.com/watch?v=2HUOyBb9FfM}{Video Link} & Romanian \\
83 & \href{https://www.youtube.com/watch?v=UAq4e6srsDQ}{Video Link} & Hindi & 113 & \href{https://www.youtube.com/watch?v=Oc6vklOiDoE}{Video Link} & Romanian \\
84 & \href{https://www.youtube.com/watch?v=qSdCSp7hGL0}{Video Link} & Hindi & 114 & \href{https://www.youtube.com/watch?v=jtlAXROMsJE}{Video Link} & Romanian \\
85 & \href{https://youtu.be/Cwj1rRJDOwI}{Video Link} & Hindi & 115 & \href{https://www.youtube.com/watch?v=G4ZxeHeQbdA}{Video Link} & Romanian \\
86 & \href{https://youtu.be/0zCFiKIx5x0}{Video Link} & Hindi & 116 & \href{https://www.youtube.com/watch?v=kXoINEaVPYo}{Video Link} & Romanian \\
87 & \href{https://youtu.be/DyBi72AIKJQ}{Video Link} & Hindi & 117 & \href{https://www.youtube.com/watch?v=sGrhsjkvWU4}{Video Link} & Romanian \\
88 & \href{https://www.youtube.com/watch?v=lmjQ4ymSqnw}{Video Link} & Hindi & 118 & \href{https://www.youtube.com/watch?v=kLtkmmXQNZc}{Video Link} & Romanian \\
89 & \href{https://www.youtube.com/watch?v=2QmfG06ZEcQ}{Video Link} & Hindi & 119 & \href{https://www.youtube.com/watch?v=UIMfGPEo48w}{Video Link} & Romanian \\
90 & \href{https://www.youtube.com/watch?v=YcGXViwXItM}{Video Link} & Hindi & 120 & \href{https://www.youtube.com/watch?v=jHfnAxISCU0}{Video Link} & Romanian \\

\hline
\end{tabular}
\end{table*}

\subsection{Topic Selection across Languages.}

\paragraph{Topic Annotation.} To ensure accurate topic categorization, each QA pair in the dataset was manually annotated and verified by human annotators, establishing ground truth topic labels for all evaluations.

Videos are selected to cover six mentorship-related topics:
\textit{Entrepreneurship, Education, Finance, Mental Health, Personal Growth}, and \textit{Physical Health}.
Topic labels are automatically assigned using \textit{Qwen2.5-7B} \cite{yang2025qwen2.5-1m} and subsequently verified by expert annotators with proficiency in the respective languages.

\smallskip
\noindent\textbf{English.}
English videos are collected from the ACL Mentorship channel\footnote{\url{https://www.youtube.com/@aclmentorship}}.
ACL Mentorship is a Year-Round Mentorship initiative of the Association for Computational Linguistics (ACL), aimed at supporting students and early-career researchers entering the field of NLP. The channel hosts virtual mentorship sessions featuring mentors from academia and industry, covering topics such as choosing a research direction, writing papers, pursuing a PhD, and maintaining work–life balance. 
The initiative promotes equal access to career guidance and currently supports a global community of over 2,500 members.

\smallskip
\noindent\textbf{Romanian.}
Romanian videos are collected from the MindArchitect channel\footnote{\url{https://www.youtube.com/@MindArchitect}}, which features conversations with coaches, educators, athletes, and other specialists. Topics span education and career development (e.g., \textit{``How do we prepare young people for the labor market?''}), as well as personal growth and mental health (e.g., \textit{``From perfectionism and self-criticism to the joy of play and mindful creation''}).

\smallskip
\noindent\textbf{Chinese.}
The Chinese videos are collected from multiple public channels rather than a single mentorship program. We curate videos that exhibit strong mentorship characteristics across mental health, career growth, finance, and physical health. The selected content primarily consists of podcast-style conversations, discussions, and interview-based dialogues involving two or three participants, typically led by domain professionals. These discussions emphasize experience-based guidance, reflective reasoning, and practical advice drawn from real world practice. Although the videos originate from diverse creators, they are unified by their focus on mentorship-focused discourse and sustained explanatory depth.

\smallskip
\noindent\textbf{Hindi.}
Hindi videos are collected from long-form interview and podcast recordings that feature in-depth conversations with domain experts, industry practitioners, entrepreneurs, and high-performing competitive exam candidates. These interactions emphasize experiential knowledge, preparation methodologies, career decision-making, and reflective discussions on challenges and learnings. The conversational format allows speakers to elaborate on nuanced perspectives and real-world insights, making the content particularly rich for extracting mentorship-focused question–answer pairs. 

\section{Methodology}

\paragraph{Implementation Details.}
All models use the PyTorch and Hugging Face Transformers stacks, and are run in inference-only mode without any fine-tuning.  Inference is executed on a GPU node equipped with four NVIDIA L40S GPUs, each with 48 GB of memory, which allows us to host the model once and parallelize calls from different videos, segments, or agents across devices.

\subsection{Prompt Design}
\begin{figure*}[t]
    \centering
    \includegraphics[width=\textwidth]{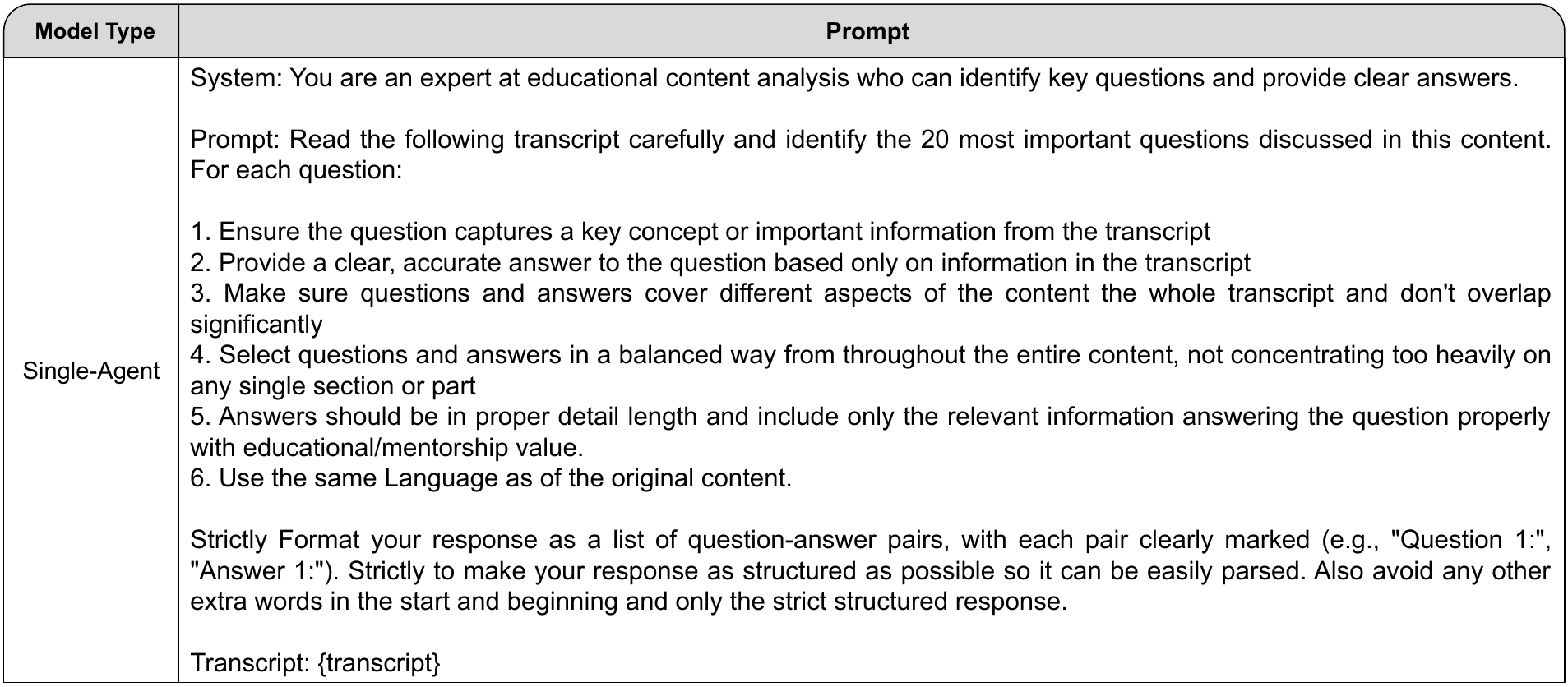}
    \caption{Single-Agent prompt.}
    \label{fig:prompt-single}
\end{figure*}

\begin{figure*}[t]
    \centering
    \includegraphics[width=\textwidth]{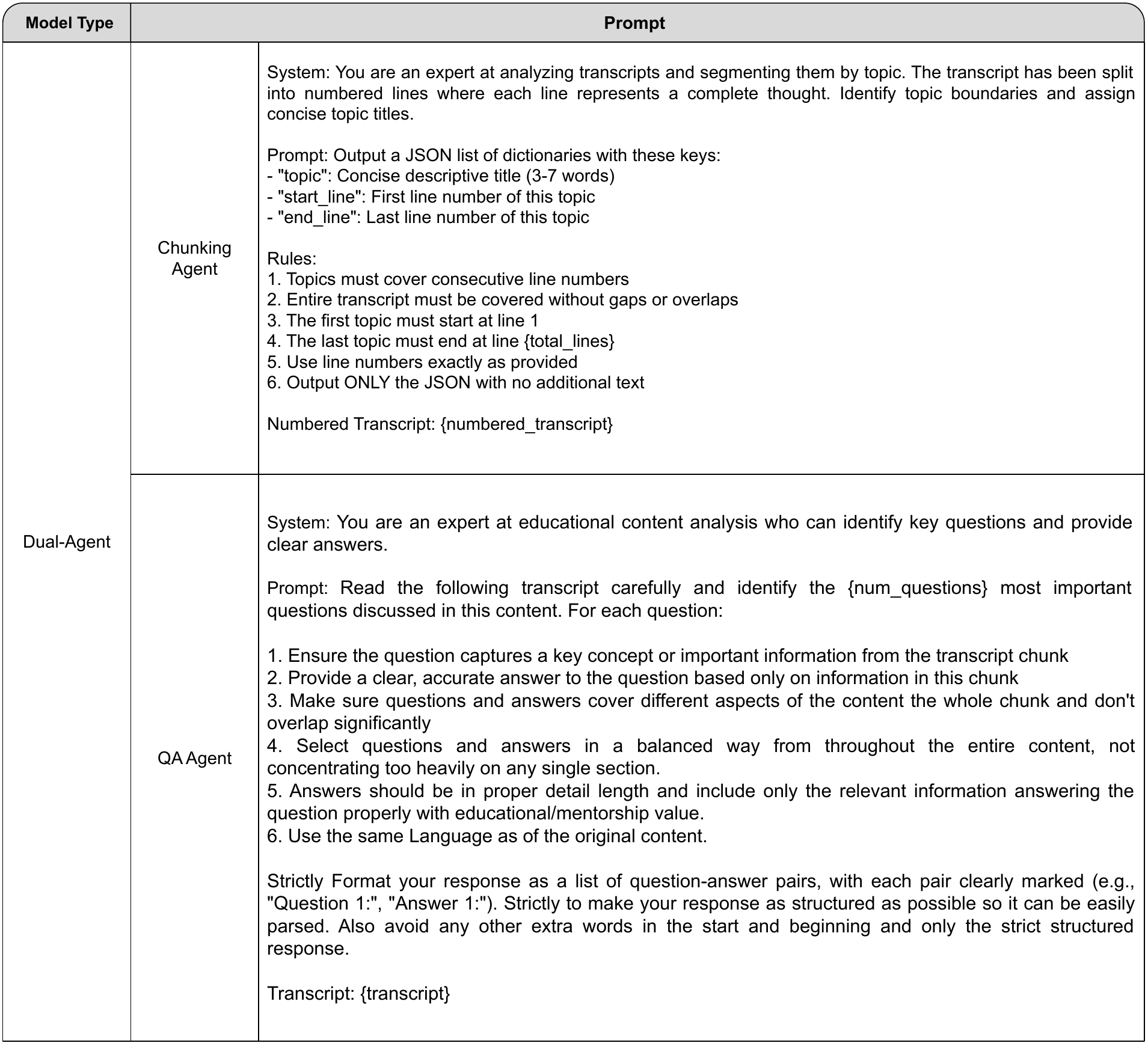}
    \caption{Dual-Agent prompt.}
    \label{fig:prompt-dual}
\end{figure*}

\begin{figure*}[t]
    \centering
    \includegraphics[width=\textwidth]{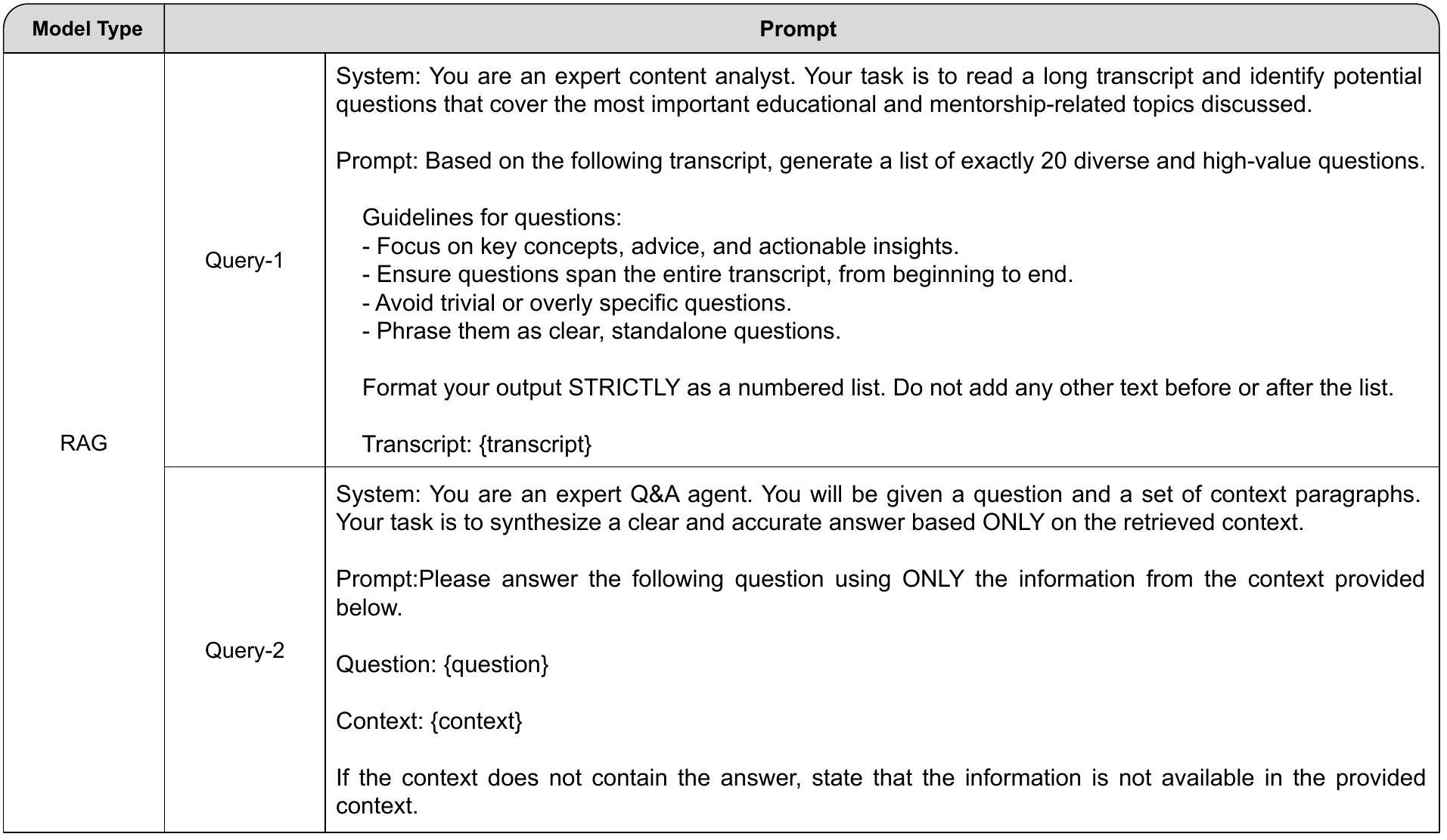}
    \caption{RAG prompt.}
    \label{fig:prompt-rag}
\end{figure*}

\begin{figure*}[t]
    \centering
    \includegraphics[width=\textwidth]{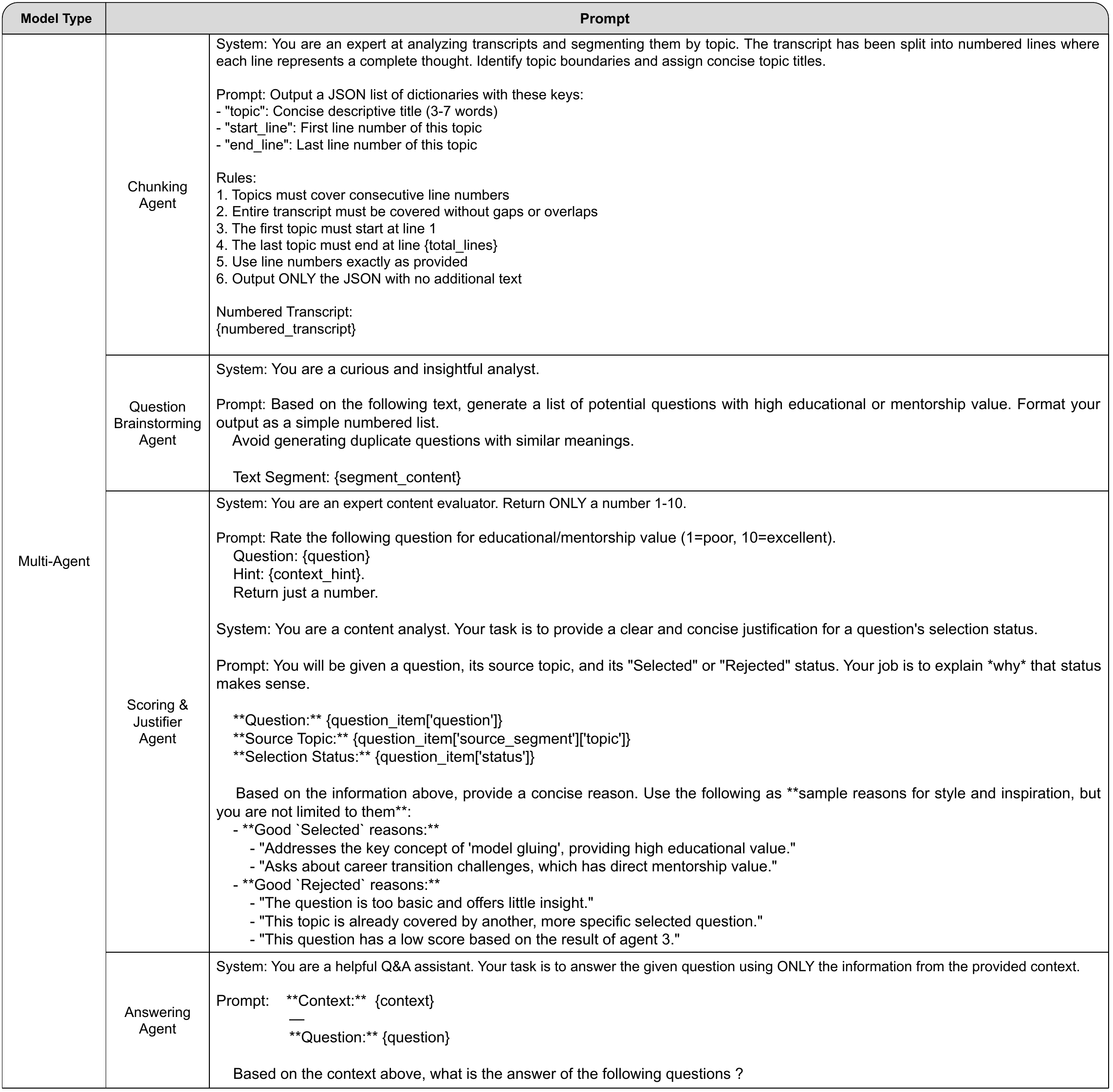}
    \caption{Multi-Agent prompt.}
    \label{fig:prompt-multi}
\end{figure*}

See \cref{fig:prompt-single}, \cref{fig:prompt-dual}, \cref{fig:prompt-rag} and \cref{fig:prompt-multi} for the prompt design.

\label{sec:prompts}

\section{Evaluation}
\label{sec:eval}

\begin{figure*}[t]
\centering
\includegraphics[width=\textwidth]{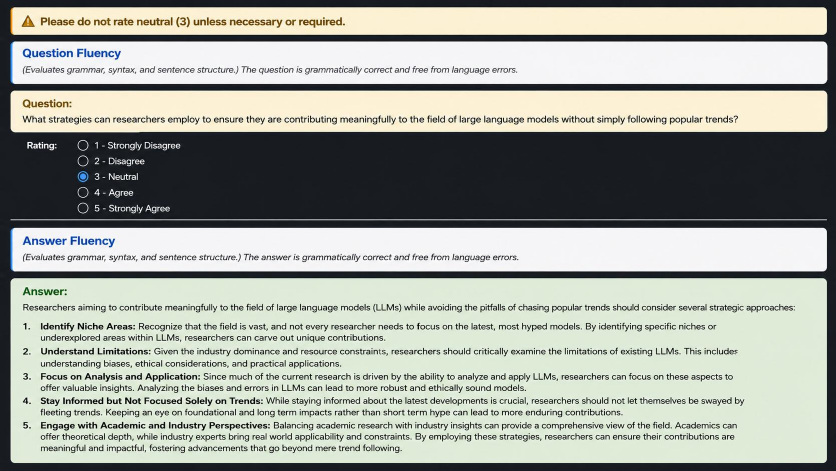}
\caption{Human evaluation interface used to rate question and answer quality on the 5 point Likert scale.}
\label{fig:Human_UI}
\end{figure*}

Figure \ref{fig:Human_UI} shows the interface used in the human annotation process.

\subsection{Can the metrics accurately assess the generated QA pairs?}

\subsubsection{Inter-Annotator Agreement}

\begin{figure}[ht]
\centering
\includegraphics[width=\columnwidth]{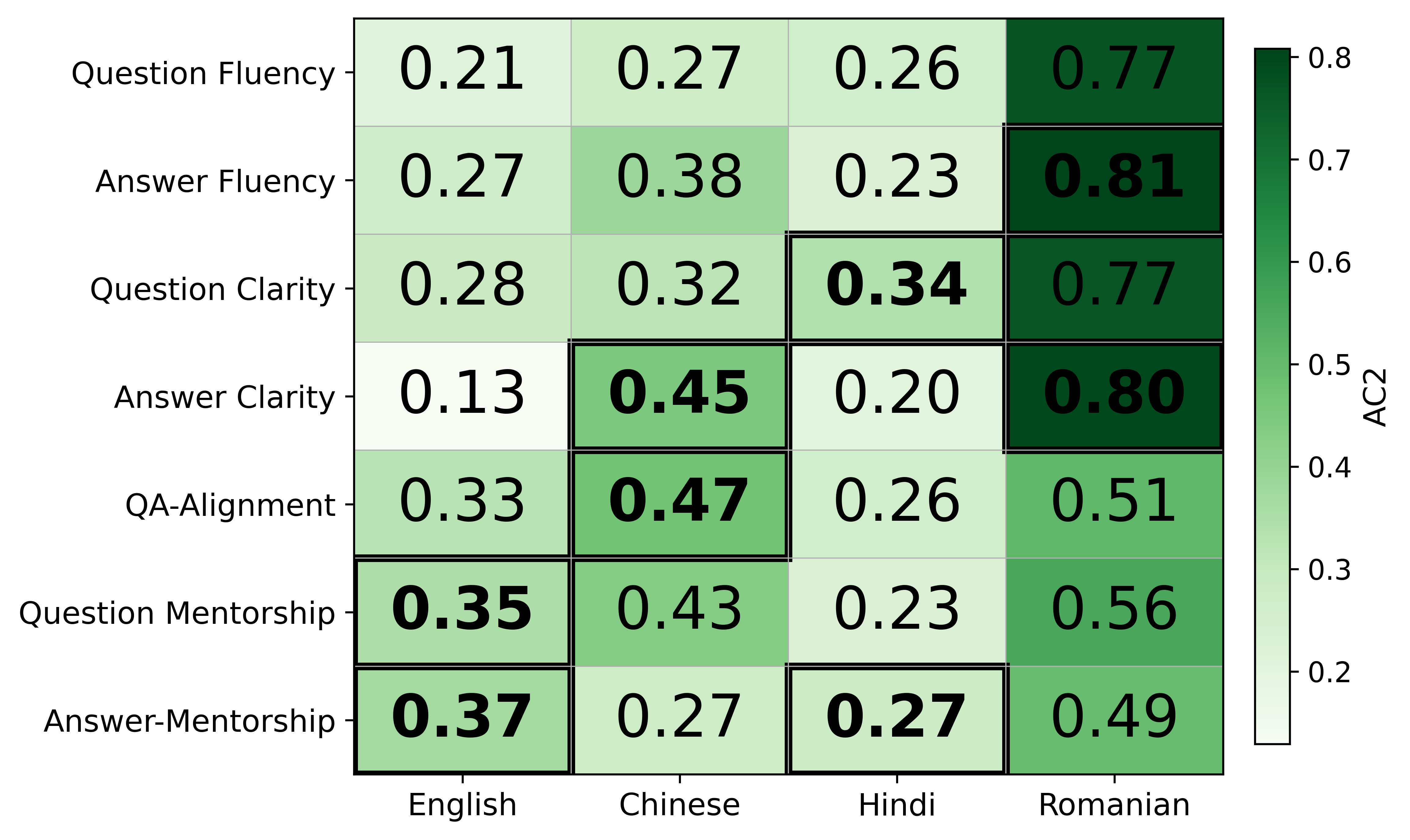}
\caption{Inter-Annotator Agreement AC2 Scores by Language}
\label{fig:inter-ann-lang}
\vspace{-0.5em}
\end{figure}

\Cref{fig:inter-ann-lang} reveals substantial variation in inter-annotator agreement across languages. \textbf{Romanian exhibits the highest and most consistent agreement across metrics} ($0.49$–$0.81$), demonstrating strong to near-perfect reliability in evaluating mentorship-focused quality. \textbf{Chinese displays moderate to fair agreement} ($0.27$–$0.47$), with metrics like \textit{Answer Clarity} and \textit{QA-Alignment} achieving reasonable inter-annotator consensus. \textbf{English and Hindi show lower but still fair agreement} ($0.13$–$0.37$ and $0.20$–$0.34$ respectively), with Hindi's challenges potentially stemming from tokenization issues and weaker model support for Devanagari script, making consistent evaluation more difficult even for human annotators. Overall, these results suggest that \textbf{annotation reliability is highly language-dependent}, reinforcing the need for culturally aware evaluation frameworks when designing mentorship-focused QA assessment protocols.

\subsubsection{LLM-Human Agreement}

\begin{figure}[ht]
\centering
\includegraphics[
    width=\columnwidth,
    trim=20 10 20 10,
    clip
]{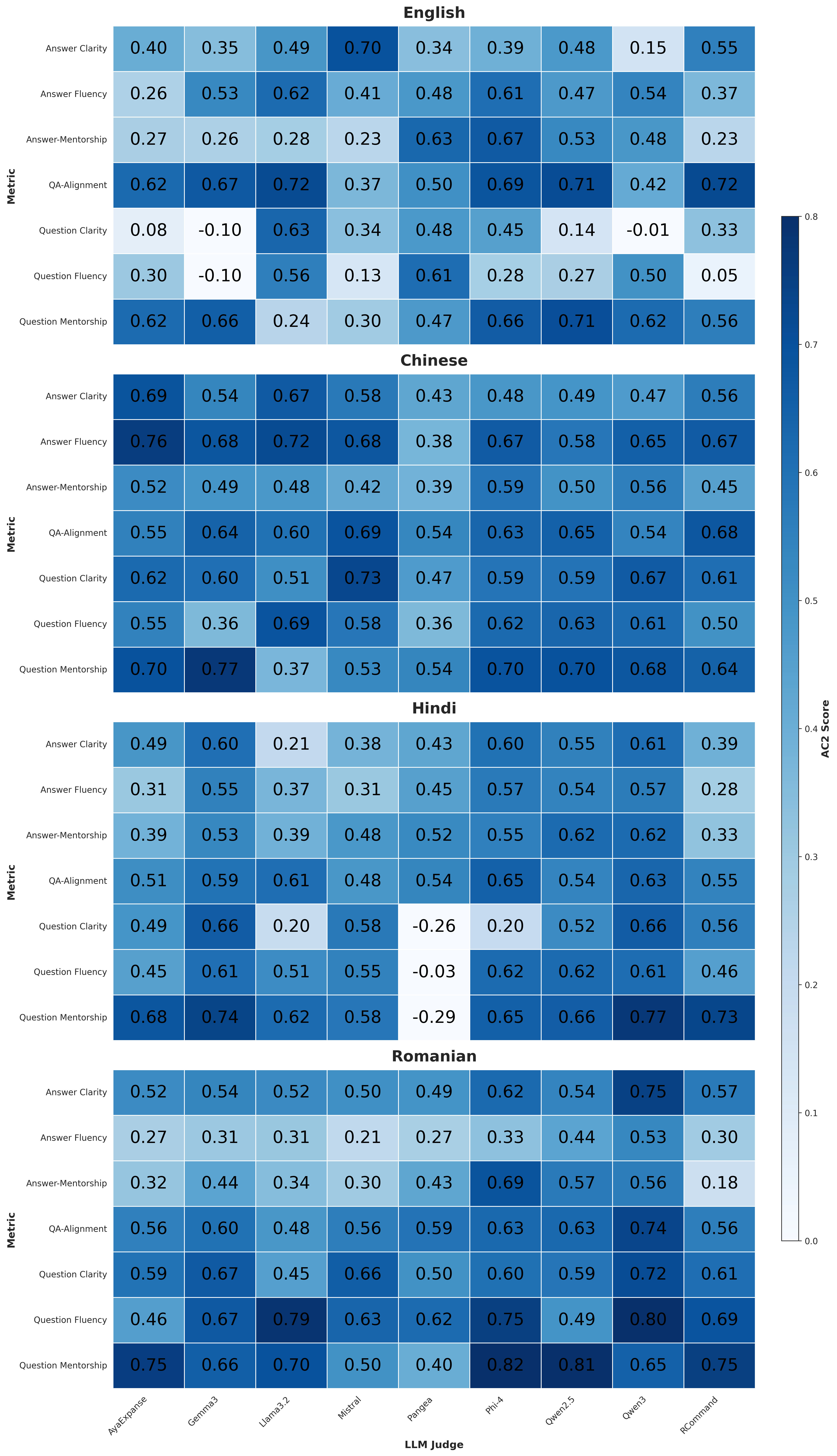}
\caption{LLM-Human AC2 Agreement Scores}
\label{fig:llm_human_all_lang}
\end{figure}

LLM Human Gwet's AC2 agreement scores per language are shown in \Cref{fig:llm_human_all_lang}.

\subsection{How do QA-generation models perform across metrics?}\label{sec:model-perf}

\begin{figure}[ht]
\centering
\includegraphics[
    width=\columnwidth,
    trim=15 8 15 8,
    clip
]{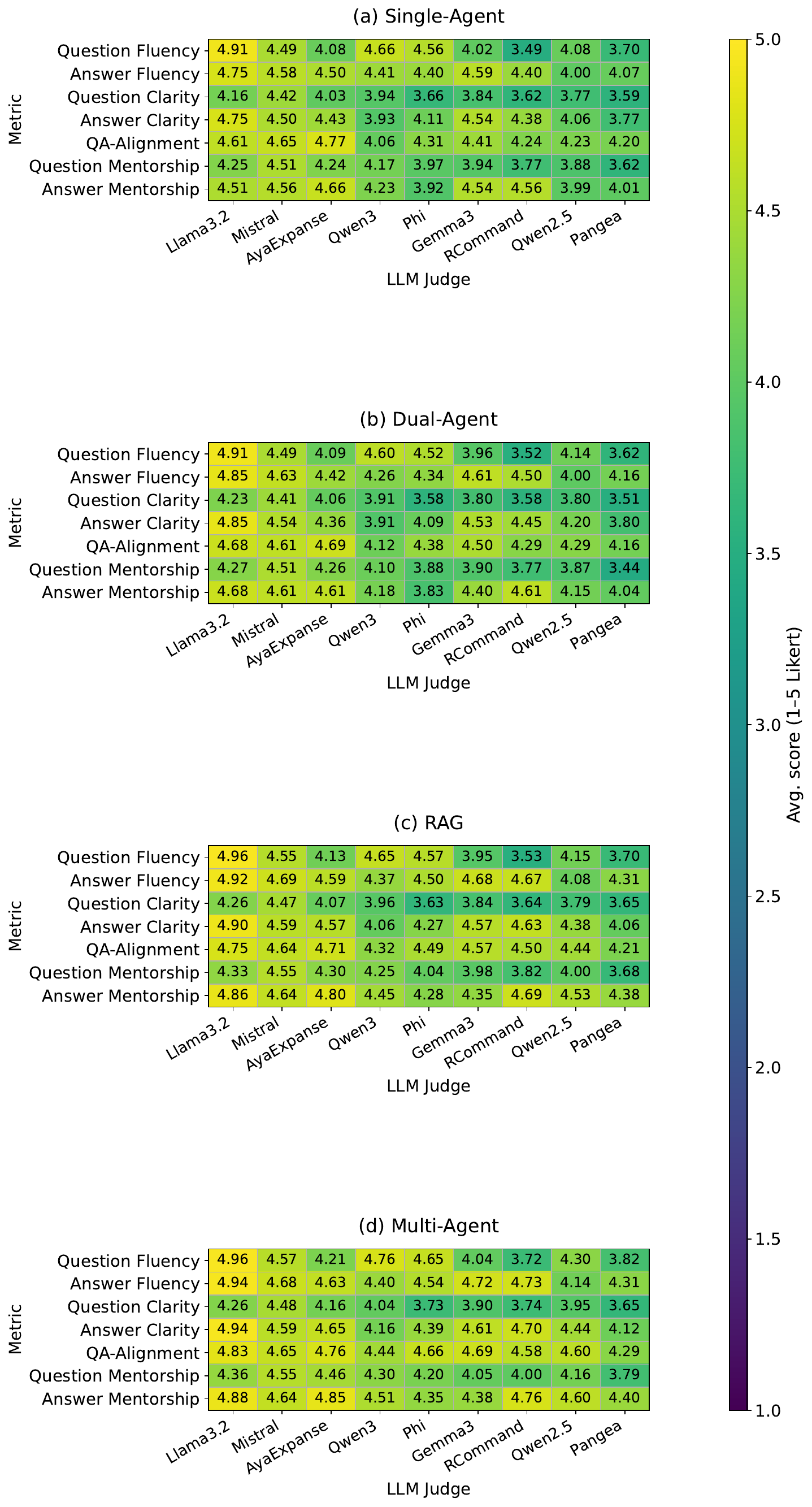}
\caption{Judge-wise performance across QA-generation models: (a) Single-Agent, (b) Dual-Agent, (c) RAG, (d) Multi-Agent. Scores are averaged over metrics using a shared 1--5 Likert scale.}
\label{fig:judge_performance}
\end{figure}

All model performance across LLM-Judges is shown in \Cref{fig:judge_performance}.

Qualitative results per model are shown below:

\paragraph{Single-Agent}

\noindent\textbf{Q:} What is the most important criterion when reviewing PhD applications?\\
\noindent\textbf{A:} Passion is the most important criterion when reviewing PhD applications.\\
\noindent\textbf{Scores:} QFlu~4, AFlu~5, QClar~5, AClar~1, QA-Algn~5, QMen~4, AMen~1

\paragraph{Dual-Agent}

\noindent\textbf{Q:} Why is it important to interact with students according to the speaker in the transcript?\\
\noindent\textbf{A:} Interacting with students is one of the most rewarding aspects of the academic path, as it allows the speaker to work with brilliant PhD students and other students across various levels, including undergraduates, masters, and postdoctoral fellows.\\
\noindent\textbf{Scores:} QFlu~5, AFlu~5, QClar~4, AClar~4, QA-Algn~5, QMen~4, AMen~3

\paragraph{RAG}

\noindent\textbf{Q:} How do PhD admission processes differ internationally?\\
\noindent\textbf{A:} PhD admission processes vary internationally in terms of interview conduct and committee structures. For instance, MIT has a stringent committee system where individual professors' admissions decisions are heavily scrutinized, whereas Stanford uses a rolling admissions process. Universities like UIUC or Hong Kong allow faculty members significant autonomy in admitting PhD students. This diversity highlights the importance of considering a broader range of institutions beyond prestigious American universities.\\
\noindent\textbf{Scores:} QFlu~5, AFlu~5, QClar~4, AClar~3, QA-Algn~5, QMen~5, AMen~3

\paragraph{Multi-Agent}

\noindent\textbf{Q:} What differences exist between PhD admission processes in various academic institutions?\\
\noindent\textbf{A:} "The differences between PhD admission processes in various academic institutions primarily revolve around the structure of the selection committees, the role of individual faculty members, and the criteria used for evaluating applications. For instance:

1. **Committee Structure**: At some institutions, like MIT, there is a strict committee process where individual professors cannot independently admit students without consensus from the committee. This contrasts with universities like UIUC or Hong Kong, where faculty members have more autonomy to select students based on their own preferences.

2. **Faculty Autonomy**: In some systems, such as those in the U.S., faculty members play a significant role in admitting PhD students. They may have the authority to interview and accept candidates directly, which can vary widely depending on the institution and department.

3. **Application Review Process**: The number of applications reviewed by each committee member differs. Some may skim through a large volume of applications, while others may focus on specific candidates they have been contacted by. For example, one panelist mentioned receiving numerous unsolicited emails, indicating that some professors may not read every application thoroughly.

4. **Criteria for Selection**: While the core process involves evaluating academic records, letters of recommendation, and research proposals, the emphasis on these factors can differ. For example, a close match in research interests might be more beneficial for some institutions, especially smaller programs, compared to larger committees that consider a broader range of criteria.

Understanding these differences can help PhD aspirants tailor their applications and outreach strategies to align better with the specific requirements and preferences of the institutions they are targeting."\\
\noindent\textbf{Scores:} QFlu~5, AFlu~5, QClar~5, AClar~5, QA-Algn~5, QMen~5, AMen~5

\subsection{How does QA-generation performance vary across languages and mentorship topics?}

\begin{figure}[ht]
\centering
\includegraphics[
    width=\columnwidth,
    trim=15 8 15 8,
    clip
]{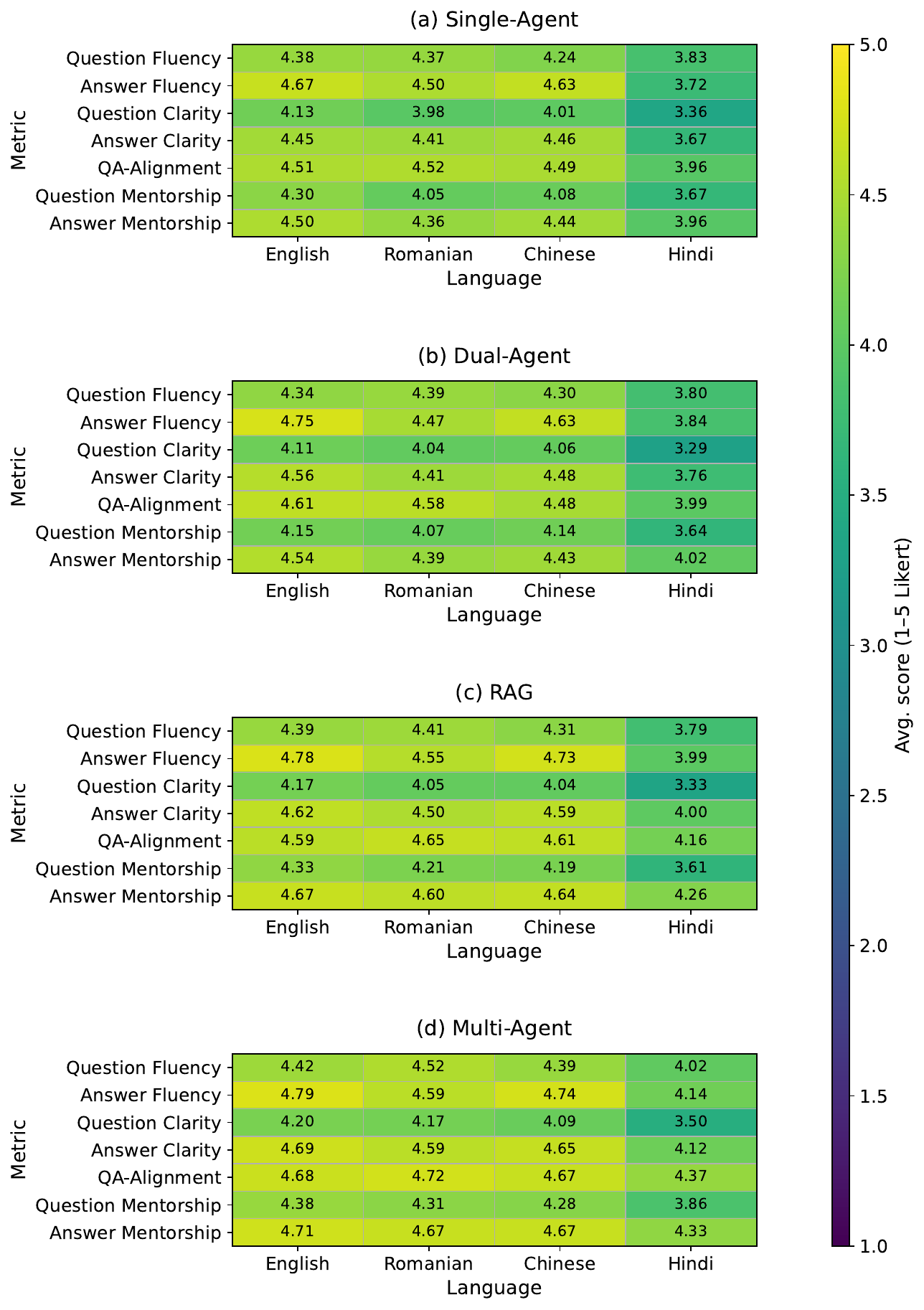}
\caption{Language-wise performance across QA-generation models: (a) Single-Agent, (b) Dual-Agent, (c) RAG, (d) Multi-Agent. Scores are averaged over human and LLM judges (1–5 Likert).}
\label{fig:all_models_language}
\end{figure}

\paragraph{Language Evaluation.}
As shown in \Cref{fig:all_models_language}, English demonstrates the strongest performance across all metrics, which is expected given its prevalence in training data. Romanian and Chinese follow closely, performing comparably to English on critical dimensions like QA-Alignment and Mentorship metrics. However, Hindi exhibits a notable performance drop across all evaluated metrics.
This disparity can be attributed to architectural and data constraints in the Qwen2.5 model. The model employs a Byte-Level Byte Pair Encoding (BBPE) tokenizer that is specifically optimized for English and Chinese \cite{yang2025qwen2.5-1m}. When processing Hindi text in Devanagari script, the tokenizer fails to recognize complete word units and instead fragments individual words into numerous byte-level tokens. This suboptimal tokenization, combined with limited Hindi representation in the training corpus, directly impacts the model's ability \cite{qwen_token} to generate and evaluate Hindi content effectively.
This interpretation is further supported by the model's reported MMLU benchmark scores: English achieves 74.37 while Hindi scores only 52.16 \cite{qwen25_14b_hindi}, reflecting a substantial performance gap. The combination of tokenization inefficiency and insufficient training exposure explains the observed quality degradation in Hindi-generated QA pairs.

The Multi-Agent model demonstrates a consistent performance advantage, effectively raising the ceiling for every language compared to the other three architectures. This improvement is evident across both high- and low-resource languages. A particularly striking improvement is observed for Hindi, which reveals the Multi-Agent model's role as a performance equalizer. In the Single-Agent architecture, Hindi substantially underperforms compared to Chinese, English, and Romanian, with QA-Alignment dropping to 3.96 and Answer Fluency falling to 3.72—both failing to reach the 4.0 threshold. However, the Multi-Agent model significantly boosts these scores, elevating QA-Alignment to 4.37 and Answer Fluency to 4.14.
This finding suggests that even when a base LLM exhibits inherent weaknesses in a specific language, implementing a Multi-Agent architecture can effectively compensate for these limitations. The approach acts as a performance equalizer, enabling the model to deliver quality that approaches or matches that of high-resource languages, thereby bypassing the model's native capability constraints.

\begin{figure}[ht]
\centering
\includegraphics[
    width=\columnwidth,
    trim=15 8 15 8,
    clip
]{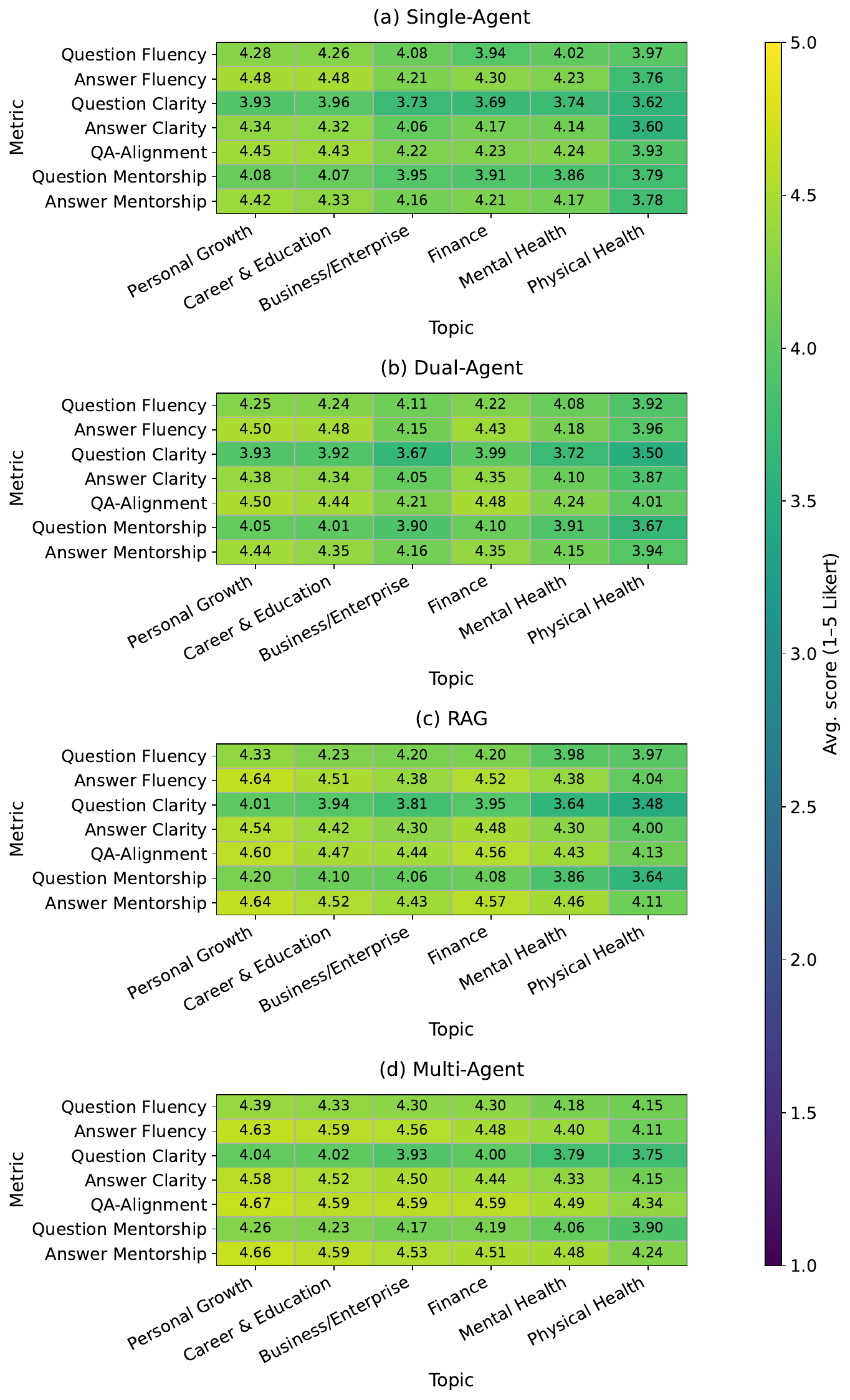}
\caption{Topic-wise performance across QA-generation models: (a) Single-Agent, (b) Dual-Agent, (c) RAG, and (d) Multi-Agent. Scores are averaged over human and LLM judges (1–5 Likert scale).}
\label{fig:all_models_topic}
\end{figure}

\paragraph{Topic Evaluation.}
As illustrated in \autoref{fig:all_models_topic}, topic complexity substantially impacts LLM judge performance across content domains. To ensure accurate topic categorization, each QA pair in the dataset was manually annotated and verified by human annotators, establishing ground truth topic labels for all evaluations.
\textbf{Personal Growth.} This domain achieves the highest scores in both Mentorship (4.55) and Alignment (4.56). The prevalence of self-help content and motivational literature in pre-training corpora enables models to generate well-structured guidance with relative ease, as the domain relies on broadly applicable principles rather than specialized knowledge.
\textbf{Physical Health.} This domain presents the greatest challenge, registering the lowest scores in Mentorship (4.04) and Alignment (4.12). Qualitative analysis reveals that speakers frequently provided medical-style guidance incorporating complex anatomical terminology (e.g., hippocampus, amygdala, prefrontal cortex), biochemical nomenclature (e.g., cortisol, dopamine, serotonin), bodily fluids, and pharmaceutical references. Models demonstrated insufficient precision for assessing such nuanced clinical content, frequently producing vague or overly generalized evaluations.
\textbf{Finance.} Despite having the smallest dataset (287 samples), Finance achieved a robust Alignment score of 4.47, comparable to more straightforward topics. This success stems from the structured, rule-based nature of financial discourse. Speakers typically provided logically structured guidance adhering to consistent frameworks (e.g., profit-loss analysis, investment strategies), enabling accurate LLM evaluation even with limited training examples.

As demonstrated in \autoref{fig:all_models_topic}, architectural complexity significantly impacts performance on challenging domains while showing minimal differentiation on straightforward topics.
\textbf{Physical Health.} This domain exhibits the most pronounced performance gap. Single Agent achieves Answer Mentorship of 3.78 and QA Alignment of 3.95, with marginal improvement in Dual Agent (3.94 and 4.01). Multi Agent substantially elevates these to 4.24 and 4.34, demonstrating clear advantages in handling clinically complex content.
\textbf{Finance.} This logic intensive domain reveals similar architectural differentiation. Single Agent produces Answer Mentorship of 4.21, Q Mentorship of 3.86, QA Alignment of 4.23, and Answer Clarity of 4.23. Multi Agent markedly improves these to 4.51, 4.20, 4.60, and 4.50, respectively.
\textbf{Personal Growth and Career.} These domains show minimal performance variation across architectures, with Single Agent achieving scores comparable to Multi Agent. This reflects the abundance of motivational content in training data, enabling simpler architectures to generate effective responses.
These results reveal domain dependent performance differentiation: common topics (Career, Personal Growth) show architectural equivalence, while challenging domains (Physical Health, Finance) demonstrate significant separation—Single and Dual Agent models degrade substantially, whereas Multi Agent maintains robust performance.

\end{document}